\documentclass[lettersize,journal]{IEEEtran}
\usepackage{amsmath,amsfonts}
\usepackage{array}
\usepackage[caption=false,font=normalsize,labelfont=sf,textfont=sf]{subfig}
\usepackage{textcomp}
\usepackage{stfloats}
\usepackage{url}
\usepackage{verbatim}
\usepackage{graphicx}
\usepackage{cite}
\usepackage{times}
\usepackage{soul}
\usepackage{url}
\usepackage[hidelinks]{hyperref}
\usepackage[utf8]{inputenc}
\usepackage[small]{caption}
\usepackage{graphicx}
\usepackage{amsmath}
\usepackage{amsthm}
\usepackage{booktabs}
\usepackage{algorithmicx}  
\usepackage{amsfonts}
\usepackage{xcolor}
\usepackage{amssymb}
\usepackage{arydshln}

\usepackage{url}

\newcommand{\tabincell}[2]{\begin{tabular}{@{}#1@{}}#2\end{tabular}}
\usepackage{multirow}

\usepackage[ruled,vlined]{algorithm2e}
\usepackage{algpseudocode}

\begin{document}

\title{BiFSMNv2: Pushing Binary Neural Networks for Keyword Spotting to Real-Network Performance}

\author{Haotong~Qin, Xudong~Ma, Yifu~Ding, Xiaoyang~Li, Yang~Zhang,\\Zejun~Ma, Jiakai~Wang, Jie~Luo, Xianglong~Liu*
\thanks{
Haotong Qin, Xudong Ma, Yifu Ding, Jie Luo, and Xianglong Liu (*corresponding author) are with Beihang University, China (email: xlliu@buaa.edu.cn).
Xiaoyang Li, Yang Zhang, and Zejun Ma are with ByteDance Inc., China.
Jiakai Wang is with Zhongguancun Laboratory, China.
}}

\markboth{Journal of \LaTeX\ Class Files,~Vol.~14, No.~8, August~2021}%
{Shell \MakeLowercase{\textit{et al.}}: A Sample Article Using IEEEtran.cls for IEEE Journals}


\maketitle

\begin{abstract}
Deep neural networks, such as the Deep-FSMN, have been widely studied for keyword spotting (KWS) applications while suffering expensive computation and storage.
Therefore, network compression technologies like binarization are studied to deploy KWS models on edge.
In this paper, we present a strong yet efficient binary neural network for KWS, namely \textbf{BiFSMNv2}, pushing it to the real-network accuracy performance.
First, we present a \textit{Dual-scale Thinnable 1-bit-Architecture} to recover the representation capability of the binarized computation units by dual-scale activation binarization and liberate the speedup potential from an overall architecture perspective.
Second, we also construct a \textit{Frequency Independent Distillation} scheme for KWS binarization-aware training, which distills the high and low-frequency components independently to mitigate the information mismatch between full-precision and binarized representations.
{Moreover, we propose the \textit{Learning Propagation Binarizer}, a general and efficient binarizer that enables the forward and backward propagation of binary KWS networks to be continuously improved through learning.
We implement and deploy the BiFSMNv2 on ARMv8 real-world hardware with a novel \textit{Fast Bitwise Computation Kernel}, which is proposed to fully utilize registers and increase instruction throughput.
Comprehensive experiments show our BiFSMNv2 outperforms existing binary networks for KWS by convincing margins across different datasets and achieves comparable accuracy with the full-precision networks (only a tiny 1.51\% drop on Speech Commands V1-12).} We highlight that benefiting from the compact architecture and optimized hardware kernel, BiFSMNv2 can achieve an impressive 25.1$\times$ speedup and 20.2$\times$ storage-saving on edge hardware.
\end{abstract}


\section{Introduction}

{Recently, deep learning techniques utilizing neural networks have achieved significant progress and have shown great potential in various fields, including but not limited to computer vision, natural language processing, speech recognition, etc~\cite{VeryDeepConvolutional,he2016deep,wang2022vpu,wang2021sequential,wang2020mead,zhang2021diversifying,guo2020multi,guo2021jointpruning,guo20223d,zhang2022motiondiffuse,DBLP:conf/cvpr/ZhaoZXLP22,DBLP:journals/tcsv/ZhaoXZLZL22,DBLP:conf/ijcai/ZhaoXZLZL20,DBLP:journals/sigpro/ZhaoXZLZ20,DBLP:conf/icmcs/00010XSHL021,DBLP:conf/cvpr/Xu0ZSL021,zhao2022cddfuse,ma2020comparisons,ma2021dirichlet,ma2019insights,ma2019fine,ma2018variational,wu2021sequential,yin2021improving}.
The deep speech processing models also emerge and become popular~\cite{ren2019fastspeech,yu2014histogram,yu2016effect,yu2017adversarial,yu2017dnn}, such as the Feedforward Sequential Memory Networks (FSMN)~\cite{zhang2015feedforward}, Compact Feedforward Sequential Memory Networks (cFSMN)~\cite{chen2018compact}, and Deep Feedforward Sequential Memory Networks (Deep-FSMN)~\cite{zhang2018deep}.
Among these, deep neural networks for keyword spotting (KWS) are increasingly being studied for real-world applications~\cite{chen2014small,zhang2015feedforward,ren2020fastspeech,berg2021keyword,leroy2019federated,yu2017spoofing}. They allow users to activate devices by speaking keywords or specific phrases while keeping the device in a dormant state when inactive.
However, these deep KWS networks are often computationally and storage expensive and are deployed on edge devices with limited resources that must constantly be on standby for possible speech input.}
Thus, the deployment of the advanced KWS models faces the significant challenge of resource constraint in practice.
To address the challenge, various novel algorithms, such as cFSMN~\cite{chen2018compact}, DC-CNN~\cite{zhang2017hello}, and BC-ResNet~\cite{kim2021broadcasted}, have been designed for lightweight deep learning for KWS. 
Though existing KWS methods have achieved remarkable storage compression and computation saving, they still suffer massive floating-point parameters and computation. Thus room for compression still exists from a bit-width perspective. 

Network binarization, as demonstrated in recent studies such as~\cite{wang2020bidet,electronics8060661,wang2022toward,zhang2021modulated,ye2021distillation,zhong2012sensitivity,qin2022distribution}, is a highly efficient form of network quantization that uses 1-bit binarized parameters to compress networks for extreme computational and storage efficiency. 
The resulting binarized models take advantage of compact weights and activations, which utilize efficient bitwise XNOR and pop-count instructions with minimal memory to save computation resources and time far more than floating-point ones. For example,~\cite{XNOR} shows that the compression and speedup ratios of binarized convolution units can theoretically reach up to 32.00$\times$ and 62.27$\times$, respectively.
However, despite the progress made in network binarization, applying existing methods to quantize neural networks for keyword spotting (KWS) still results in a significant drop in accuracy. It is because both weights and activations are binarized, which constrains the representation capability of the 1-bit models to the limited value space of the parameters. Additionally, binarization makes it challenging to optimize the models, as it introduces high discretization into deep models.
{In comparison to other compression methods such as pruning and architecture design, network binarization possesses powerful topological characteristics as it only affects parameters~\cite{qin2023bibench,guo2021jointpruning,sandler2018mobilenetv2}. And it is widely researched in academic studies as a standalone compression technique rather than simply a 1-bit specialization of quantization~\cite{HWGQ,TTQ}.}
Furthermore, existing architectures for KWS have fixed model scales and topologies, which cannot adaptively balance resource budgets at runtime. Moreover, the existing deployment frameworks are still far from reaching the theoretical upper limit of acceleration for binarized networks when implemented on real-world hardware.

\begin{figure*}[!ht]
\centering
\includegraphics[width=18cm]{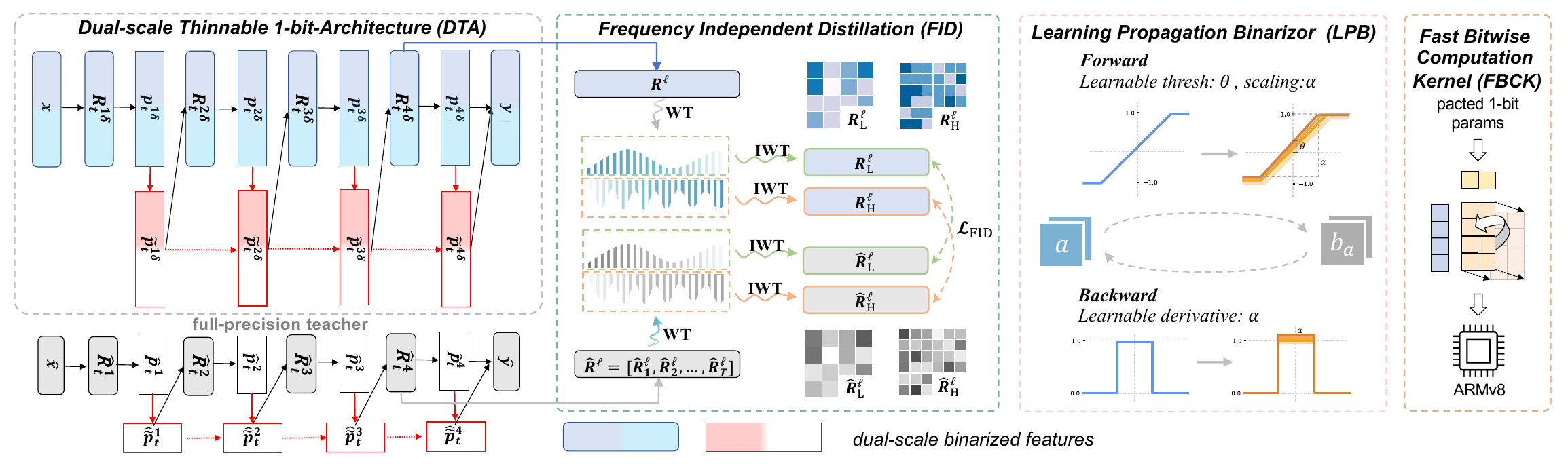}
\caption{{Overview of BiFSMNv2. The binarization framework includes (1) Dual-scale Thinnable 1-bit-Architecture (DTA) to strengthen the representation of the binarized computing units and balance accuracy-efficiency trade-off, (2) Frequency Independent Distillation (FID) to distill the high and low-frequency components independently from the full-precision network, (3) Learning Propagation Binarizer to improve the forward and backward propagation of binary KWS networks continuously through learning, and (4) Fast Bitwise Computation Kernel (FBCK) to 
achieve the full potential of binarized models on real-world devices.}} 
\label{fig:overview}
\end{figure*}

This paper presents a lightweight and strong binary neural network for the KWS task dubbed as \textbf{BiFSMNv2}, with lightweight computation yet high accuracy (see the overview in Fig.~\ref{fig:overview}).
BiFSMNv2 is constructed based on the binarization of Deep-FSMN with a pure feedforward structure.
(1) We present a \textit{Dual-scale Thinnable 1-bit-Architecture} (DTA), which recovers the representation capability of the binarized computing units by dual-scale activation binarization and liberates the speedup potential from an overall architecture perspective.
(2) We also construct a \textit{Frequency Independent Distillation} (FID) scheme for training binarized KWS models, which distills the high and low-frequency components independently to match corresponding information between the full-precision and binarized representations.
{(3) We design the \textit{Learning Propagation Binarizer} (LPB), a general and efficient binarizer that enables the forward and backward propagation of binary KWS networks to be continuously improved through learning.}
(4) BiFSMNv2 is implemented and deployed on real-world ARMv8 devices with an efficient \textit{Fast Bitwise Computation Kernel} (FBCK), which enjoys impressively faster inference than that using the existing binarization frameworks.

This work extends our prior conference publication~\cite{qin2022bifsmn}, which presented the first binary network specifically for keyword spotting (KWS) tasks. Despite efforts made in the field, existing binary networks still have a significant accuracy gap compared to their full-precision counterparts due to high discretization. To address this, we further improve the optimization, representation, and architecture of binarization in this work and compare the proposed method with state-of-the-art (SOTA) binarization algorithms.
{Compared to the conference version, the detailed improvements of this work are presented as follows:}
\begin{itemize}
\item {We improve the architecture for the binarized KWS network, which increases the performance of binarized computing units while maintaining computational flexibility during inference. Based on the adaptively thinnable architecture, we further recover the binarized representation through dual-scale feature construction and liberate the speedup potential through an adaptively thinnable architecture from an overall perspective.}
\item {We present a stronger distillation scheme for binarized KWS networks that independently distills the high and low-frequency components from the full-precision network, enabling binarization-aware distillation to mitigate the mismatch of information capacity from high and low frequencies and fully utilize the knowledge from full-precision representations.}
\item {We propose a generic yet efficient binarizer that enables the forward and backward propagation of binary KWS networks to be continuously improved through learning. It introduces a learnable threshold to mean-shift the representation in forward and a ratio parameter to achieve gradient error reduction in backward.}
\item {We compare our BiFSMNv2 with more SOTA binarization algorithms on keyword spotting, including XNOR++\cite{XNOR++}, ReActNet\cite{liu2020reactnet}, ReCU~\cite{xu2021recu}, and FDA~\cite{xu2021learning}. Moreover, we compare it with more binarized architectures (including a compact BiFSMNv2$_\text{S}$ version) in Table~\ref{tab:architectures}. The results show that BiFSMNv2 performs well among various binarization algorithms and architectures on different KWS datasets.}
\item {Benefiting from the latest implementation, we greatly reduce the number of parameters and inference calculations while improving the accuracy and achieving 20.2$\times$ compression and 25.1$\times$ acceleration in the deployment of ARMv8 hardware.}
\end{itemize}

Our BiFSMNv2 first pushes binary neural networks for KWS to real-network accuracy.
Comprehensive experiments on Google Speech Commands V1 and V2 datasets~\cite{warden2018speech} show that BiFSMNv2 completely outperforms state-of-the-art (SOTA) binarization methods and is even almost lossless compared with full-precision counterparts among 12, 20, and 35 keyword classification tasks, {\textit{e.g.}, BiFSMNv2 enjoys 96.34\% accuracy on Speech Commands V1-12 with a tiny 1.51\% drop.}
Besides, we highlight that the efficient implementation makes our BiFSMNv2 easy to deploy and fast to inference in real-world devices on edge ARM devices. BiFSMNv2 can achieve up to 20.2$\times$ storage-saving and 25.1$\times$ speedup compared with the full-precision Deep-FSMN. The code is released at \textcolor{blue}{\url{https://github.com/htqin/BiFSMNv2}}.

\section{Related Work}

\subsection{Neural Network Binarization}
\label{sec:NetworkBinarization}
With various advanced binarization methods emerging, the neural networks can be compressed and accelerated by binarization despite the different neural architectures or downstream learning tasks.~\cite{qin2022distribution,jiang2022toward,li2022local}. 
There are several ways to generically improve binarized networks. The typical practices are (1) quantization error minimization: reducing the information loss during sign function transformation from 32-bit value to 1-bit value~\cite{XNOR}; (2) loss function improvement: optimizing the binarized networks to close the accuracy gap from their full-precision counterparts~\cite{Regularize-act-distribution}, or distilling the knowledge from representations of full-precision networks to binarized networks~\cite{RealtoBin}; (3) gradient approximation: designing the approximation of binarization function for the backward propagation to match the gradient~\cite{Regularize-act-distribution}.
Moreover, a practice to boost the performance for different architectures is to design the specific binarization algorithm or binarized structure for each network and task, such as CNNs~\cite{BiReal,liu2020reactnet}, transformer-based~\cite{BinaryBERT,qin2022bibert} and MLP-based networks~\cite{qin2020bipointnet} for computer vision, natural language processing, \textit{etc}. But as for KWS tasks, the general binarization algorithms are not satisfying due to the high accuracy demands, and even the specific binary neural network suffers a significant accuracy gap.
Binary neural networks nowadays heavily depend on the support of hardware libraries when deployed. The binarization operations are constructed by XNOR and pop-count instructions on target hardware platforms (\textit{e.g.}, CPUs, GPUs, and FPGAs). Though some binarization frameworks, including daBNN~\cite{zhang2019dabnn} and Bort~\cite{bolt}, can deploy some representative networks and tasks, such as ResNets for image classification, the support is still not ready for KWS networks.

\subsection{Deep Neural Networks for Keyword Spotting}

Deep neural networks are widespread in many applications, including KWS tasks, due to their overwhelming superiority in performance. 
The recurrent neural network (RNN)~\cite{9414339} is originally designed to capture the context in a sequence, combining the advantages of convolutional and recurrent layers to utilize both the local structure and long-range context. However, RNNs are usually computationally and energy expensive with poor parallelism on hardware. 
To reduce the consumption, CNN-based (BC-ResNet~\cite{kim2021broadcasted}) models for KWS are also proposed with better performance and less energy cost, which applies most residual functions as 1D temporal convolution while still allowing 2D convolution together to expand temporal output to frequency-temporal dimension.
Besides, RNNs also suffer from the vanishing of gradient, which is solved by Feedforward Sequential Memory Networks~\cite{zhang2015feedforward} afterward. FSMN has many variants, such as compact Feedforward Sequential Memory Networks (cFSMN)~\cite{chen2018compact}, Deep Feedforward Sequential Memory Networks (Deep-FSMN)~\cite{zhang2018deep}, and pyramidal Feedforward Sequential Memory Networks (pFSMN)~\cite{yang2018novel}. They are efficient in computation and also fast in convergence. 
However, FSMNs still conduct floating-point operations, which have redundancy in parameters and leave room for model quantization and binarization to further compress and accelerate inference on resource-constraint devices.

\section{BiFSMNv2}

We first introduce a basic binarization framework on Deep-FSMN for KWS application and then present BiFSMN equipped with \textit{Frequency Independent Distillation} (FID) and \textit{Dual-scale Thinnable 1-bit-Architecture} (DTA) techniques, and efficient \textit{Fast Bitwise Computation Kernel} (FBCK) for deployment on ARMv8 devices. 

\subsection{Basic Binarization Framework}
\label{subsec:BinarizedDeep-FSMNArchitecture}
Firstly, we build a basic binarization framework on Deep-FSMN~\cite{zhang2018deep} architecture for the KWS task. Deep-FSMN is one of the variants of FSMN, which is particularly deep with shortcuts to capture inherent features in the input speech audios. It is binarization-friendly since (1) Deep-FSMN is a pure feed-forward structure stacked by memory blocks to capture and store the information of sequential input, (2) the skip connections in adjacent layers are proved important for the accuracy performance of binarized networks \cite{BiReal} to allow the information to flow directly to the next block.

In the typical binarization network, we replace the floating-point parameters (weights and activations) with 1-bit by inserting $\operatorname{sign}$ function in the forward propagation:
\begin{equation}
\label{eq:sign_function}
\boldsymbol{b_x}=\operatorname{sign}(\boldsymbol{x})=
\begin{cases}
1& \text{if } \boldsymbol{x} \ge 0\\
-1& \text{otherwise }
\end{cases}.
\end{equation}
{And as for the backward propagation, the Straight-Through-Estimator (STE)~\cite{courbariaux2015binaryconnect} is a simple but widespread solution for the problem of zero-gradient:}
\begin{equation}
\label{eq:backward_gx}
\boldsymbol{g_x}=
\begin{cases}
\boldsymbol{g}_{\operatorname{sign}(\boldsymbol x)}& \text{if } |\boldsymbol x| \leq 1\\
0 & \text{otherwise }
\end{cases},
\end{equation}
{where $\boldsymbol{g_x}$ and $\boldsymbol{g}_{\operatorname{sign}(\boldsymbol x)}$ denote the gradients of $\boldsymbol x$ and $\operatorname{sign}(\boldsymbol x)$, respectively. $\boldsymbol x$ denotes the element in floating-point weights or activations.}
The scaling factor $\alpha$ is only for weights to keep the magnitude of the real values:
\begin{equation}
\label{eq:bwn}
\alpha_{\boldsymbol{w}} = \operatorname{mean}(|\boldsymbol{w}|),
\end{equation}
where $\operatorname{mean}(\cdot)$ denotes taking the mean value.
We make $\boldsymbol{w} \approx \alpha_{\boldsymbol{w}} \boldsymbol{b_w}$ to reduce the quantization error, where $\boldsymbol{b_w}$ is the 1-bit weight binarized by $\operatorname{sign}$ function.
Therefore, the whole computation process of the convolutional layer is denoted as follows:
\begin{equation}
\label{eq:bwn}
\boldsymbol{o} = \alpha_{\boldsymbol{w}} \left(\boldsymbol{b_w}\otimes\boldsymbol{b_a}\right),
\end{equation}
where $\boldsymbol{b_w}$, $\boldsymbol{b_a}$, and $\boldsymbol{o}$ denote binarized weights, binarized activations, and outputs, respectively, and $\otimes$ denotes the inner product with bitwise operation constructed by XNOR and pop-count instructions.

Applying the binarization operation on both the floating-point weights and activations for linear and convolutional layers, where the $\operatorname{Matmul}$ operation costs the majority of the computational resources and time, we build the basic binarized Deep-FSMN. We denote the input $\ell$-th hidden states as $\boldsymbol{R}^\ell=[\boldsymbol{R}_1^\ell, \boldsymbol{R}_2^\ell, ..., \boldsymbol{R}_T^\ell]$, and for each $\boldsymbol{R}_t^\ell, t\in[1, T]$,  which is the fixed-size representation of the long context surrounding at time instance $t$. The $\ell$-th binarized memory block in the network can be written as: 
\begin{equation}
\begin{aligned}
\tilde{\boldsymbol{p}}_{t}^{\ell}=&\sum_{i=0}^{N_{1}^{\ell}} \alpha_{\boldsymbol{a}_i}^\ell\left(\boldsymbol{b}_{\boldsymbol{a}_{i}}^{\ell} \otimes \boldsymbol{b_p}_{t-is_{1}}^{\ell}\right)\\
+&\sum_{j=1}^{N_{2}^{\ell}} \alpha_{\boldsymbol{c}_j}^\ell\left(\boldsymbol{b}_{\boldsymbol{c}_{j}}^{\ell} \otimes \boldsymbol{b_p}_{t+js_{2}}^{\ell}\right)\\
+&\mathcal{H}\left(\tilde{\boldsymbol{p}}_{t}^{\ell-1}\right)+\boldsymbol{p}_{t}^{\ell}.
\end{aligned}
\end{equation}
where {$\mathcal{H}(\cdot)$ means the skip connection (identity mapping) within the memory block.}
The output of the binarized linear projection layer is denoted as
\begin{equation}
{\boldsymbol{p}}_{t}^{\ell}=\alpha_{\boldsymbol{V}}\left({\boldsymbol{b_V}}^{\ell}\otimes{\boldsymbol{b_h}}_{t}^{\ell}\right)+{\boldsymbol{b}}^{\ell},
\end{equation}
while $\tilde{\boldsymbol{p}}_{t}^{\ell}$ denotes the output of the full-precision memory block. $N_{1}^{\ell}$ and $N_{2}^{\ell}$ are the look-back and lookahead orders and $s_1$ and $s_2$ are the strides for look-back and lookahead filters respectively. 
The next hidden layer is consists of $T$ time instances as $\boldsymbol{R}^{\ell+1}=[\boldsymbol{R}_1^{\ell+1}, \boldsymbol{R}_2^{\ell+1}, \cdots, \boldsymbol{R}_T^{\ell+1}]$, each of the time instance $t$ can be expressed as:
\begin{equation}
\label{eq:4}
\hat{\boldsymbol{R}}_{}^{\ell+1}=f\left(\alpha_{\boldsymbol{U}}^{\ell}\left(\boldsymbol{b}_{\boldsymbol{U}}^{\ell}\otimes \tilde{\boldsymbol{b}}_{\boldsymbol{p}_t}^{\ell}\right)+\boldsymbol{b}^{\ell+1}\right),
\end{equation}
where $f(\cdot)=\operatorname{BN}\cdot\operatorname{Nonlinear}(\cdot)$ denotes the composition of batch normalization and nonlinear functions (PReLU in the binarized network~\cite{RealtoBin}).

\begin{figure}[t]
\centering
\includegraphics[width=7cm]{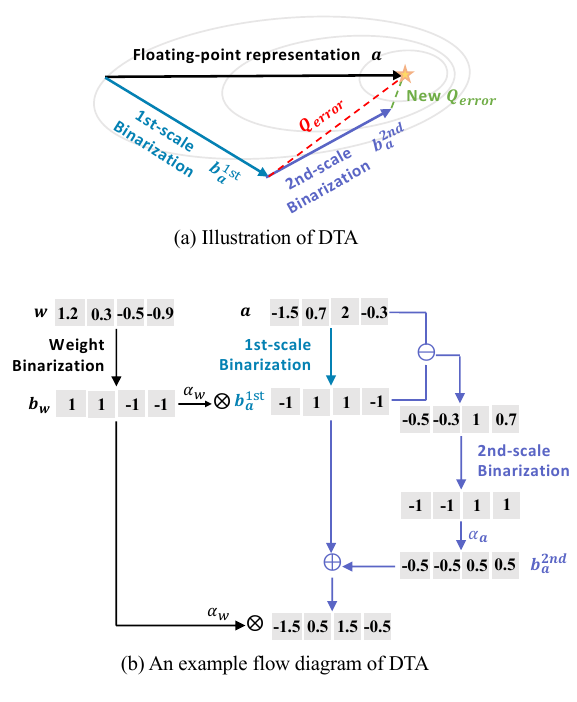}
\caption{The illustration and example of DTA for a single computing unit. (a) The DTA applies a dual-scale binarization to activations, aiming to recover their representation capability by reducing the quantization error. {(b) We present an example flow diagram for the activation binarization of DTA.}}
\label{fig:MRB}
\end{figure}

\subsection{Dual-scale Thinnable 1-bit-Architecture: Adapting Runtime Resources for Accuracy-efficiency Trade-off}

As discussed earlier, binarization is an efficient compression approach with extremely lightweight 1-bit parameters and efficient bitwise operations, enabling fast inferences on resource-limited devices. 
However, there are still two challenges in the practical use of binary neural networks:
{(1) Feature binarizing causes severe degeneration of representation capability in the prediction.}
(2) The energy budget varies between devices and is even different in wake-up and power-saving modes. 
Therefore, an adaptive yet powerful binarized architecture is required to permit instant accuracy-efficiency trade-offs for KWS applications in runtime.

\begin{figure}
\centering
\includegraphics[width=9cm]{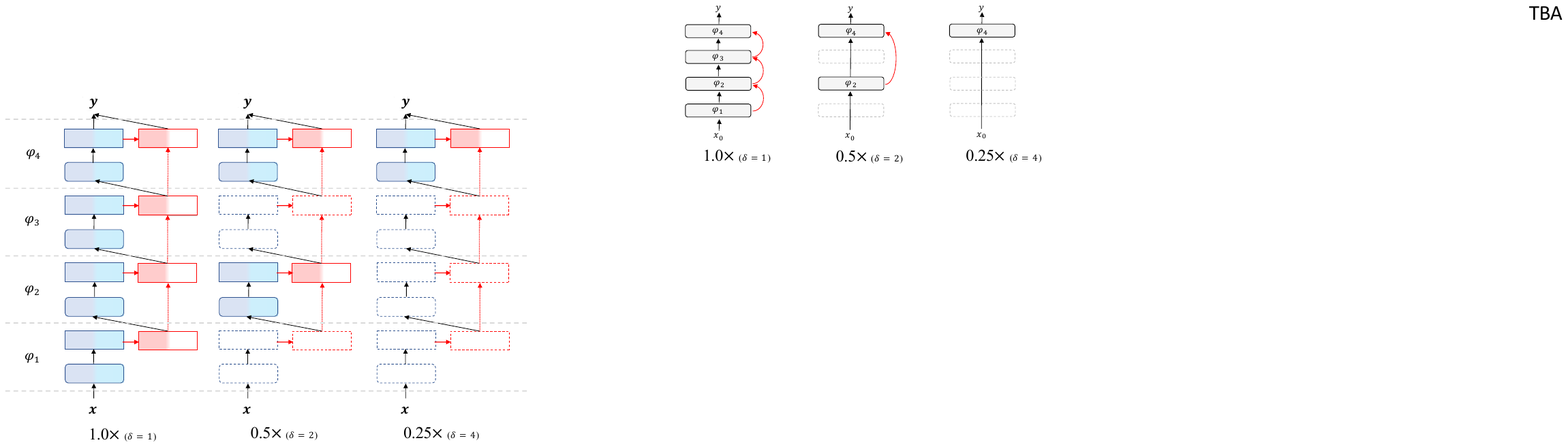}
\caption{An instance of Dual-scale Thinnable 1-bit-Architecture ($N=4$, $\delta=1, 2, 4$) at runtime, where red arrows denote skip connections.}
\label{fig:thinnable}
\end{figure}

{In existing keyword spotting networks that have been proposed, including full-precision and binary neural networks~\cite{zhang2015feedforward,zhang2018deep,chen2018compact,qin2022bifsmn}, we can disassemble their design motivations of the network architectures into two steps: first, defining individual layers and/or blocks, such as using residuals connections and 1-bit computationally-intensive layers; then, define the combination layers and/or blocks, including thinnable architectures and block-stacking structures. One of the well-recognized challenges for binary neural networks is that the representation capability of a single binarization layer is severely limited due to its restricted weights and activations. Therefore, it is promising to improve the trade-off in the architecture design specifying for the binary KWS network, sacrificing the computational savings of a single 1-bit layer/block to allow stronger representation capabilities and cooperate with a lighter overall topology.}

{Along with this motivation, we present a Dual-scale Thinnable 1-bit-Architecture (DTA) for KWS. In the proposed architecture, the representation capability of binarized computing units is significantly strengthened while the overall topology is simplified to be lighter; thus, it enjoys a balance between accuracy and efficiency and enables switching to thinner versions at runtime for online resource constraints.
First, we build the strengthened binarized computing units to recover their representation capability, including 1-dim convolutional, 2-dim convolutional, and linear units.}
As Eq.~(\ref{eq:bwn}) presents, the activation is directly binarized by $\operatorname{sign}$ function, which makes representations highly discrete and loses information. We introduce a dual-scale binarization to quantizing computing units, where the first and second scales of features are
\begin{equation}
\begin{aligned}
&\textrm{1st scale:}\quad\boldsymbol{b}_{\boldsymbol{a}}^\textrm{1st}=\operatorname{sign}(\boldsymbol{a}),\\
&\textrm{2nd scale:}\quad\boldsymbol{b}_{\boldsymbol{a}}^\textrm{2nd}=\alpha_{\boldsymbol{a}}^\textrm{2nd} \operatorname{sign}(\boldsymbol{a}-\boldsymbol{b}_{\boldsymbol{a}}^\textrm{1st}),
\end{aligned}
\end{equation}
where $\alpha_{\boldsymbol{a}}^\textrm{2nd}=\operatorname{mean}(|\boldsymbol{a}-\boldsymbol{b}_{\boldsymbol{a}}^\textrm{1st}|)$ is the scaling factor of the second-scale residual feature and should be calculated in real-time during inference. Since the more fine-grained feature is introduced, the single binarized unit can represent more detailed information. Taking the convolutional unit as an example, the process can be formulated as
\begin{equation}
\label{eq:dual}
\boldsymbol{o}_\textrm{dual} = \alpha_{\boldsymbol{w}} \left(\boldsymbol{b_w}\otimes\boldsymbol{b}_{\boldsymbol{a}}^\textrm{1st}\right)+\alpha_{\boldsymbol{w}}\alpha_{\boldsymbol{a}}\left(\boldsymbol{b_w}\otimes\boldsymbol{b}_{\boldsymbol{a}}^\textrm{2nd}\right),
\end{equation}
and the binarized linear unit has the similar form. The illustration and example of DTA are presented in Fig.~\ref{fig:MRB}.

Our dual-scale binarization strengthens binarized representations, focusing on recovering the fine-grained binarized activations. 
{In previous literature, a few works also recognized this positive effect on the accuracy of binarized and applied multiple or fine-grained weights/activations to boost representation~\cite{XNOR,DBLP:journals/corr/abs-1708-08687,ABCNet}.} 
Different from these studies, we solve representation degeneration of binarization directly with minimal cost. First, we find that the capability of the binarized model is mainly limited by binarized activations, which reflects that the information degradation of the data flow is more severe than the weights with scaling factors. Thus, we do not introduce additional weights and maintain the storage consumption of binarized computing units the same as before. Moreover, since scaling factors for activations would be calculated in real-time during inference, we only use scaling factors on the second-scale binarized features instead of re-scaling for two activations. It adds little computational overheads while greatly recovers the information loss of the purely binarized one (only first-scale). {However, although our dual-scale binarization can significantly enhance representation capabilities with limited storage and computation costs, applying it directly to the original overall topology still brings additional computational burden (compared to vanilla binarization methods). 
For example, for an 8-layer Deep-FSMN architecture, the inference FLOPs of directly using dual-scale quantization are about $1.5\times$ that of applying the DoReFa-Net binarization method~\cite{dorefa}.
The strengthened binarized units request us to design a lighter overall architecture.}

{Therefore, since the strengthened binarized computing units with dual-scale binarization bring additional computation overhead, we should also design a flexible and compact architecture for an accuracy-efficiency trade-off.
When we focus on the full-precision backbone in FSMN, which is computationally expensive, the basic binarization architecture $\textrm{M}$ containing $2N$ blocks ($2N$ defaults to 8) is expressed as:}
\begin{equation}
\textrm{M}(\boldsymbol x) = \varphi^{2N}\cdot\varphi^{2N-1}\cdot...\cdot\varphi^1(\boldsymbol x),
\end{equation}
where $\textrm{M}$ and $\varphi^\ell$ are the binary neural network and $\ell$-th binarized Deep-FSMN block, respectively, $\boldsymbol x$ is the input of network.
A compact thinnable binarization architecture $\textrm{M}_\textrm{DTA}$ using dual-scale binarized computing units can be derived from $\textrm{M}$ for adaptive switching on deployment:
\begin{equation}
\textrm{M}_\textrm{DTA}(\boldsymbol x; \delta) = \Phi_\textrm{dual}^{N}\cdot\Phi_\textrm{dual}^{N-1}\cdot...\cdot\Phi_\textrm{dual}^1(\boldsymbol x),
\end{equation}
where the block number $N$ is reduced to $N$ due to the strengthened single-unit representation capability, and $\delta$ is the interval of selected layers, which is confined to be divisible into $N$.
And each thinnable block ${\Phi_\textrm{dual}^{\ell}}$ can be defined as
\begin{equation}
{\Phi_\textrm{dual}^{\ell}}(\boldsymbol x)=
\begin{cases}
\varphi_\textrm{dual}^{\ell}(\boldsymbol x), & \ell\in \{i\delta, i \in[1, N/\delta]\},\\
\boldsymbol x, & \textrm{otherwise}.
\end{cases}
\end{equation}
And the batch normalization in the function $f(\cdot)$ for $\ell$-th dual-scale binarized block ${\varphi^\ell_\textrm{dual}}$ are preassigned as Eq.~(\ref{eq:dual}) according to different variants in the thinnable network.
The thinnable network architecture will skip intermediate blocks in every $\delta$ layer by replacing them with identity functions. 

{Fig.~\ref{fig:overview} shows the formalization of our proposed dual-scale thinnable 1-bit-architecture, and we also provide an instance for the $N=4$, $\delta=1, 2, 4$ setting in Fig.~\ref{fig:thinnable}, which is also our default experiment setting.
Under this setting, compared with the existing binary KWS network, our architecture has stronger representation capability and lower resource consumption.}

\subsection{Frequency Independent Distillation: Matching Information in Binarization-aware Training}

\begin{figure*}[t]
\centering
\includegraphics[width=\linewidth]{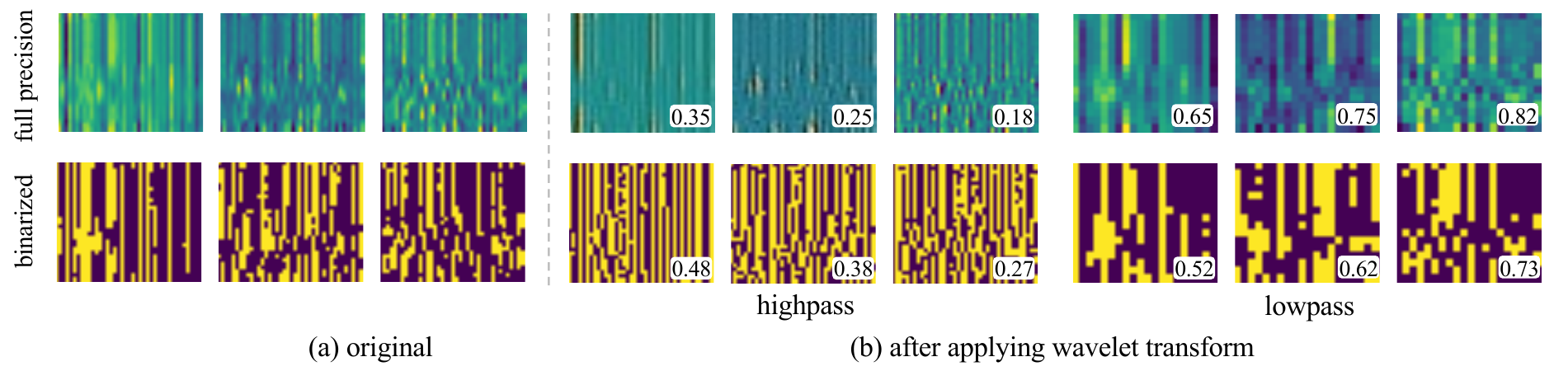}
\caption{Visualization of intermediate representations for some samples between the full-precision network and the binarized one. Values in the bottom-right are the relative wavelet energy of the high-/low-frequency components. The high-frequency component is much larger than the low-frequency one in binarized representation.}
\label{fig:wavelet}
\end{figure*}

Usually, the accuracy of binarized networks will drop severely compared with full-precision counterparts since the representation capability is extremely limited by compact parameters. We visualize the intermediate representation of both full-precision and binarized networks in Fig.~\ref{fig:wavelet}. It is a natural result that the information is richer in the full-precision but monotonous in the binarized ones since it restricts the values to binary. Therefore, some binarization methods are devoted to globally retaining the information using the binarized representations during training. For example, \cite{qin2019forward} bases on entropy-maximization and \cite{liu2020reactnet} applies learnable mean-shift to both weights and activations. 
However, in binarized representations, the outlines of the object may convey more important information than plane blocks. The values on the edge always fall to 1 or -1, which displays the main feature of the input. While compared to the plane blocks, the edges are the locally gradient-maximized part of the feature map and are hard to be directly optimized globally.

Fortunately, we find that the information trends to gather in edges because the binarized features are better at capturing high-frequency components. 
First, we define the high and low-frequency components of the intermediate representations and the related metrics.
We use 2D Haar Wavelet Transform (WT)~\cite{meyer1992wavelets} to isolate the horizontal and vertical edges as one of the most frequently used separable transforms. By this way, the intermediate representations are decomposed into low and high-frequency components. 
We sum up the wavelet function family weighted, and the inputted representation $\boldsymbol{R}$ of one specific layer can be written as: 
\begin{equation}
\label{eq:haar}
f_\textrm{WT}(\boldsymbol{R}) = \sum^{-1}_{j=-N} \sum_k \boldsymbol{C}_j(k) \phi_{j,k},
\end{equation}
where $\phi$ is the mother wavelet function with a specific time parameter, $j=-1, \cdots, -N$ is the resolution level, and $k$ determines the translation of the waveform.
The relative wavelet energy is used as a metric to measure the information amount of the intermediate representation that has been decomposed:
\begin{equation}
\label{eq:wavelet-energy}
\mathcal{E}_j = \sum_k | \boldsymbol{C}_j(k) |^2.
\end{equation}
Therefore, the relative wavelet energy of low and high-frequency components ($\boldsymbol{C}_j\in\{\boldsymbol{C}_\textrm{L}, \boldsymbol{C}_\textrm{H}\}$, $N=2$) after once decomposition could be represented as:
\begin{equation}
p_\textrm{H}=\frac{\mathcal{E}_\textrm{H}}{\mathcal{E}_\textrm{total}},\qquad p_\textrm{L}=\frac{\mathcal{E}_\textrm{L}}{\mathcal{E}_\textrm{total}}.
\end{equation}
where $\mathcal{E}_\textrm{total}=\mathcal{E}_\textrm{H}+\mathcal{E}_\textrm{L}$ denotes the total relative wavelet energy of representation.
If the information is richer in some components, the corresponding energy is also larger. 

Then we can obviously find the differences in frequency for the representations of full-precision and binarized networks.
As Fig.~\ref{fig:wavelet} shows, compared with the representation in the full-precision network, the relative wavelet energy (values in the bottom-right) of the high-frequency components rapidly increases after being binarized but the low-frequencies decreases, which indicates that the binarized representation inclines higher-frequency components.
As mentioned in Section~\ref{sec:NetworkBinarization}, many current binarization-aware training methods focus on making binarized features close to full-precision ones, including numerical approximating and distillation-based approaches. Our findings suggest that directly learning from full-precision to binarized features will lead to a mismatch between their high and low-frequency components. Specifically, comparing the full-precision to binarized representations directly is equal to consider not only the content of high and low-frequency but also the proportion of frequency information. Since the capability of capturing frequency is fundamentally different for the two representations, the mismatch caused by the proportion gap will usually exist and be hard to mitigate, which in turn significantly affects training performance.

According to the discussion above, we design a distillation scheme that is tailored to binarization-aware training, named Frequency Independent Distillation (FID). 
As general distillation schemes, a well-trained full-precision Deep-FSMN model is required as the teacher model, and the FID scheme is applied to transform knowledge from a full-precision model to a binarized one.
Specifically, we decompose the original representations by Haar Wavelet Transform (WT) for full-precision teacher and binarized student models to make the frequency components of their representations independent and then obtain the representations of corresponding components by applying Inverse Wavelet Transform (IWT).
For the teacher model, the process can be formulated as follow:
\begin{equation}
\begin{aligned}
\hat{\boldsymbol{R}}_{\textrm{H}} &= f_\textrm{IWT}\left(\sum_k \hat{\boldsymbol{C}}_{\textrm{H}}(k) \phi_{T\textrm{H},k}\right),\\
\hat{\boldsymbol{R}}_{\textrm{L}} &= f_\textrm{IWT}\left(\sum_k \hat{\boldsymbol{C}}_{\textrm{L}}(k) \phi_{T\textrm{L},k}\right),
\end{aligned}
\end{equation}
where $\hat{\boldsymbol{C}}_{\textrm{H}}(k)$ and $\hat{\boldsymbol{C}}_{\textrm{L}}(k)$ denote the high and low-frequency components, respectively, and $\hat{\boldsymbol{R}}_{\textrm{H}}$ and $\hat{\boldsymbol{R}}_{\textrm{L}}$ denote the corresponding transformed representations. And for the student model, we can also obtain $\boldsymbol{R}_{\textrm{H}}$ and $\boldsymbol{R}_{\textrm{L}}$ following the same steps.
Then, inspired by \cite{RealtoBin}, we minimize the distillation loss with attention form between separated representations from the teacher and student models, which is expressed as
\begin{equation}
\label{eq:loss-r2b}
\mathcal{L}_\textrm{FID} = \sum^{N}_{\ell=1} \left\| \frac{\boldsymbol{R}^{\ell\ 2}_{\textrm{H}}}{\left\|\boldsymbol{R}^{\ell\ 2}_{\textrm{H}}\right\|} - \frac{\hat{\boldsymbol{R}}^{\ell\ 2}_{\textrm{H}}}{\left\| \hat{\boldsymbol{R}}^{\ell\ 2}_{\textrm{H}}\right\|} \right\|+\left\| \frac{\boldsymbol{R}^{\ell\ 2}_{\textrm{L}}}{\left\|\boldsymbol{R}^{\ell\ 2}_{\textrm{L}}\right\|} - \frac{\hat{\boldsymbol{R}}^{\ell\ 2}_{\textrm{L}}}{\left\| \hat{\boldsymbol{R}}^{\ell\ 2}_{\textrm{L}}\right\|} \right\|,
\end{equation}
where $\ell$ denotes the $\ell$-th block and $\|\cdot\|$ is the L2-norm.
Since the frequency mismatch between binarized and full-precision representations is mitigated, the proposed FID scheme makes the optimization of the binarized student network accurate by utilizing the knowledge from the full-precision model.

Furthermore, to optimize the proposed TDA, we adopt a uniform layer mapping strategy in FID to align and learn representation: 
\begin{equation}
\label{eq:distill_loss_delta}
\mathcal{L}_{\textrm{FID}}^\delta = \sum^{N/\delta}_{i=1} \left\| \frac{\boldsymbol{R}^{i\delta\ 2}_{\textrm{H}}}{\left\| \boldsymbol{R}^{i\delta\ 2}_{\textrm{H}}\right\|} - \frac{\hat{\boldsymbol{R}}^{i\delta\ 2}_{\textrm{H}}}{\left\| \hat{\boldsymbol{R}}^{i\delta\ 2}_{\textrm{H}}\right\|} \right\|
+\left\| \frac{\boldsymbol{R}^{i\delta\ 2}_{\textrm{L}}}{\left\| \boldsymbol{R}^{i\delta\ 2}_{\textrm{L}}\right\|} - \frac{\hat{\boldsymbol{R}}^{i\delta\ 2}_{\textrm{L}}}{\left\| \hat{\boldsymbol{R}}^{i\delta\ 2}_{\textrm{L}}\right\|} \right\|.
\end{equation}
The gradients from different switches are accumulated during backward propagation to update the weight jointly. According to the compression ratio in thinnable architecture, the weighted loss can be calculated as
\begin{equation}
\label{eq:total_loss}
\mathcal{L}_{\textrm{tot}} = \sum_\delta \frac{1}{2^{\delta-1}} \left(\mathcal{L}_{\textrm{CE}}^\delta + \gamma \mathcal{L}_{\textrm{FID}}^\delta \right),
\end{equation}
where $\mathcal{L}_{\textrm{CE}}^\delta$ denotes the cross-entropy loss of $\mathrm{M}_\textrm{DTA}(\cdot; \delta)$ and $\gamma$ is a hyperparameter to control distillation impact, set to 0.01 as default. 

\subsection{{Propagating Learnable Binarizer: Learning Forward and Backward Propagation for Binarization}}

{Through our DTA architecture and FID optimization design, we have developed a powerful and lightweight binary neural network that can perform accurate and flexible reasoning on edge devices. However, our research has shown that this is still not the limit of what binary neural networks can achieve on KSW tasks. There is still room for improvement in existing parameter binarizers, which could lead to even more efficient and effective binary neural networks for KWS.}

\begin{figure}[t]
\centering
\includegraphics[width=9cm]{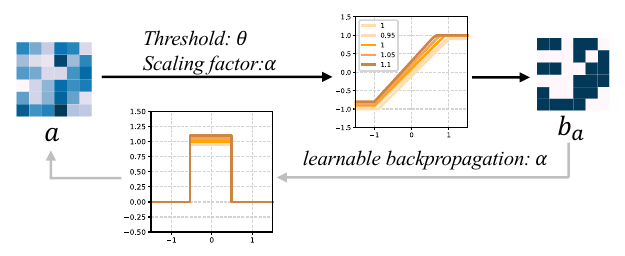}
\caption{{An illustration for the learnable propagation binarizer. We introduce a learnable threshold $\theta$ to mean-shift the representation in forward propagation and a ratio parameter $r$ of the approximate function in the backward propagation to reduce the gradient error. }}
\label{fig:MRB}
\end{figure}

{As mentioned in Section~\ref{subsec:BinarizedDeep-FSMNArchitecture}, in binary neural networks, binarizers are used to quantize the weights and activations of the computation-intensive units in the neural network. Specifically, during forward propagation, the parameters are quantized to 1 and -1 using the discrete sign function (as shown in Eq.~(\ref{eq:sign_function})); during backward propagation, the clip function is applied to approximate the derivative of the sign function in order to update the parameters (Eq.~(\ref{eq:backward_gx})). 
We find that existing binarizers have limitations in both forward and backward, impacting the expression capability of the binarized parameters significantly: }

\noindent{(1) Rigid forward parameterization. Using the sign function in the training process results in full-precision parameters being directly quantized to -1 or 1 with a threshold of 0. This makes it difficult for the quantizer to adjust to the overall distribution, as direct adjustments at the global level are challenging to achieve. The adjustment of the quantizer to the overall distribution must rely on indirect pixel-level parameter updates, which greatly restricts the representation capability of the binarized parameters.}

\noindent{(2) Backward error accumulation. The use of the clip function leads to the gradient in the region $x\in [-1, 1]$ being passed directly with a ratio of 1. However, the derivative of the sign function in this area is just 0, resulting in a ratio close to infinity, indicating a significant error that cannot be adjusted during the training process. This error can lead to suboptimal models and negatively impact the overall performance.}

\noindent{These observations suggest that currently applied binarizers have limitations in both forward and backward propagation, which can significantly affect the expression capability of the quantized parameters. Further research is needed to develop new binarizers that address these limitations, to improve the performance of binarized neural networks.}

{We propose the Learning Propagation Binarizer (LPB), a generic yet efficient binarizer that enables the forward and backward propagation of binary KWS networks to be continuously improved through learning. In the forward propagation, we introduce a learnable threshold $\theta$ based on the sign function to mean-shift the representation. Through this operation, the features of the forward propagation can be conveniently shifted from the global level, and the diversity is greatly enriched, thereby improving the representation ability of the binary quantization model. In the backward propagation, we also introduce a ratio parameter $r$ for gradient approximation and adjust the ratio of the approximate function in the backward propagation to reduce the gradient error. The forward and backward propagation of LPB can be expressed as follows:}
\begin{equation}
\label{eq:quantized}
{\textrm{Forward:}\ \operatorname{LPB}(\boldsymbol{x})=\operatorname{sign}(\boldsymbol{x}-\theta)=
\begin{cases}
1& \text{if } \boldsymbol{x} \ge \theta \\
-1& \text{otherwise }
\end{cases},}
\end{equation}
\begin{equation}
{\textrm{Backward:}\ \frac{\partial\operatorname{LPB}(\boldsymbol{x})}{\partial\boldsymbol{x}}\approx
\begin{cases}
r \boldsymbol{g}_{\boldsymbol{x}} & \text{if } |\boldsymbol{x}-\theta| \leq r\\
0 & \text{otherwise }
\end{cases},\qquad}
\end{equation}
{where $\operatorname{LPB}(\cdot)$ denote the LBP function. Since the improvements are limited to the binarizer, it can be used universally in all binarized computing units, including convolutional and linear layers.}

The detailed training procedures for the BiFSMNv2 are listed in Algorithm~\ref{alg:1}.

\begin{algorithm}[t]
    \small
    \caption{The training process of our BiFSMNv2.}
    \label{alg:1}
    \KwIn{Fixed pre-trained full-precision teacher $\textrm{M}_\textrm{FP32}$ and thinnable binarized model $\textrm{M}_\textrm{DTA}$ (BiFSMNv2) with $N$ basic binarized blocks, training iterations $T$.}
    \KwOut{Well-trained thinnable binarized model $\textrm{M}_\textrm{DTA}$}
    \For{all $t=1,2,\dots, T$}
    {
    Forward propagate $\textrm{M}_\textrm{FP}(\boldsymbol{x})$ and obtain the information-enhanced intermediate features $\hat{\mathbb{R}}_{\text{H}}=\{\hat{\boldsymbol{R}}_{\text{H}}^1, \dots, \hat{\boldsymbol{R}}_{\text{H}}^N\}$ and $\hat{\mathbb{R}}_{\text{L}}=\{ \hat{\boldsymbol{R}}_{\text{L}}^1, \dots, \hat{\boldsymbol{R}}_{\text{L}}^N\}$\;
    \For{all $\delta=1, 2, \dots, \log_2 N$}
    {
    Forward propagate $\textrm{M}_\textrm{DTA}(\boldsymbol{x}; \delta)$ and obtain the intermediate features $\mathbb{R}_{\text{H}}=\{\boldsymbol{R}_{\text{H}}^1, \dots, \boldsymbol{R}_{\text{H}}^\delta\}$ and $\mathbb{R}_{\text{L}}=\{\boldsymbol{R}_{\text{L}}^1, \dots, \boldsymbol{R}_{\text{L}}^\delta\}$\;
    Compute the distillation loss $\mathcal{L}_{\textrm{FID}}^\delta$ by Eq.~(\ref{eq:distill_loss_delta})\;
    Compute the cross-entropy loss $\mathcal{L}_{\textrm{CE}}^\delta$\;
    }
    Descend $\mathcal{L}_\textrm{tot}$ as Eq.~(\ref{eq:total_loss}) and update $\textrm{M}_\textrm{DTA}$\;
    }
    Get the well-trained BiFSMNv2 model\;
    Evaluate the BiFSMNv2 on the test dataset and get the accuracy.
\end{algorithm}

\subsection{Fast Bitwise Computation Kernel for Efficient Hardware Deployment}

The FLOPs~\cite{BiReal} of a single binarization layer is theoretically reduced by an extremely high level of $64\times$, benefiting from the binarization weights and compression to the original bit width $\frac{1}{32}$. However, when we implement and deploy the entire binary neural network on real-world hardware using existing binarization deployment frameworks, its overall inference efficiency is often significantly lower than the theoretical upper limit, such as daBNN~\cite{zhang2019dabnn} and Bolt~\cite{bolt}. One of the key bottlenecks in speedup is the use of Binarized General Matrix Multiplication (BGEMM), which is performed by bitwise XNOR and Bitcount.
In existing deployment frameworks like daBNN, the corresponding instructions on ARMv8 are \texttt{EOR}, \texttt{CNT}, \textit{etc}.
However, owing to the limit of register and instruction throughput, the efficiency of BGEMM hardware implementation is far less than that of its theoretical design.
Therefore, for efficient deployment on edge devices with limited computational resources, we further optimize the 1-bit computation with new instruction and register allocation strategy to accelerate the inference on ARMv8-A architecture which are widely used on edge devices. We dub it Fast Bitwise Computation Kernel (FBCK). 

\begin{figure}
\centering
\includegraphics[width=6.5cm]{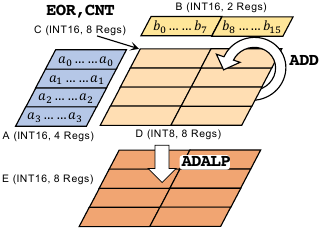}
\caption{Fast Bitwise Computation Kernel for BiFSMNv2, which improves the utilization of registers to expand instruction throughput.}
\label{fig:deploy}
\end{figure}

According to the number of registers on ARMv8 architecture, we first reallocate the registers in the kernel as five partitions in order to improve the register utilization and reduce memory footprint: partition A has four registers (except register v0) for one input (weight/activation), B has two for the other input, C has eight for intermediate results of \texttt{EOR} and \texttt{CNT}, D has eight for the output in one loop, and E has eight for the final results. 
Each input is packed as INT16. Each register in A stores an input that is repeated eight times, while each register in B stores eight different inputs. We first apply \texttt{EOR} and \texttt{CNT} for A with one register of B to get 32 INT8 results in intermediate partition C, and then perform \texttt{ADD} to accumulate the INT8 to D, and do the same for the other register of B. 
After sixteen times loop, we finally accumulate the INT8 data stored in D to an INT16 register (in E) using long instruction \texttt{ADALP}, which extends INT8 data to double width. FBCK makes full use of registers almost without idle bits during the computation. Refer to Fig.~\ref{fig:deploy} as an illustration.

\section{Experiments}
\label{sec:Experiments}

In this section, to verify the effectiveness of our BiFSMNv2, we conduct comprehensive experiments on the Google Speech Commands V1 and V2 datasets ~\cite{warden2018speech} and compare it with State-Of-The-Art (SOTA) binarization methods on various KWS deep network architectures.

\textbf{BiFSMNv2:} Our BiFSMNv2 is implemented and trained by PyTorch since it enjoys a strong automatic differentiation function and high flexibility. To build a binarized KWS model, we use binarization methods to quantize convolutional and linear units on specified neural architectures.

\textbf{Neural architecture:} We test the performance of the proposed BiFSMNv2 on both normal and compact architecture versions, dubbed as BiFSMNv2 and BiFSMNv2$_\text{S}$, respectively, on Google Speech Commands datasets in our experiments.
The BiFSMNv2 for experiments has four memory blocks with 128 backbone memory size and 224 hidden size.

To verify the accuracy and efficiency advantages of our method, we also compare BiFSMNv2 with binarized networks with different neural architectures~\cite{VeryDeepConvolutional,kim2021broadcasted,zhang2018deep,zhang2015feedforward,qin2022bifsmn}. 
We binarize the convolutional and linear units in KWS networks except for the first and last layers. 

{\textbf{Hyperparameters and training setups:}}
In order to evaluate our BiFSMNv2 fairly, we mostly apply the original hyper-parameter settings and training steps  or follow exactly the presented results in their papers~\cite{XNOR,XNOR++,BiReal,liu2020reactnet,xu2021recu,xu2021learning,qin2022bifsmn,qin2019forward}. 
Specifically, among the experiments of BiFSMNv2, we apply SGD optimizer and use a cosine annealing scheduler decaying learning rate to 0. For other networks and binarization methods, we mostly follow the hyper-parameter settings of their original papers. We train all binarized models 300 epochs, the same as their full-precision counterparts.

{\textbf{Evaluation metric definition:}
Following the existing works for KWS~\cite{kim2021broadcasted,qin2022bifsmn}, we use accuracy as the metric to evaluate the classification tasks on Speech Commands V1 and V2 datasets. The accuracy is the fraction of predictions the model got right and has the following definition:
\begin{equation}
\text{Accuracy} = \frac{\text{Number of correct predictions}}{\text{Total number of predictions}}.
\end{equation}
The metric definition is generic for 12, 20, and 35 classification tasks (CLS 12, CLS 20, and CLS 35) on Speech Commands V1 and V2 datasets.}

{\textbf{Computational FLOPs}:
For 1-bit weights and activations, the multiplication between $m$-bit and $n$-bit parameters approximately takes $mn/64$ Floating Point Operations (FLOPs) for a CPU using 64-bit instructions~\cite{dorefa,BiReal}, so we calculate the FLOPs for a single binarized computing unit as $1/64$ of its full-precision counterpart.}

{\textbf{Deployment hardware}:
To validate the practicability of BiFSMNv2, we test the actual speed of BiFSMNv2 on Raspberry Pi 3B+ with 1.2GHz 64-bit ARMv8 CPU Cortex-A53.}

\begin{table}[t]
\centering
\caption{{Ablation study of our BiFSMNv2 on Speech Command V1 (SC-V1) and V2 (SC-V2) datasets with 12 classifications.} }
\resizebox{\linewidth}{!}
{\begin{tabular}{llcrrr}
\toprule
{\bf Arch.} & {\bf Quant} & {\tabincell{c}{\textbf{\#Bits}\\$_\textrm{(W/A)}$}} & {\tabincell{r}{\textbf{FLOPs}\\$_\textrm{(M)}$}}  & {\tabincell{r}{\textbf{{SC-V1}}\\$_{(\%)}$}} & {\tabincell{r}{\textbf{{SC-V2}}\\$_{(\%)}$}}\\
\midrule
{\tabincell{l}{Deep-FSMN}} & Full Prec.    & 32/32 & 710.15 & 97.93 & 98.05 \\
\midrule
\multirow{4}{*}{\tabincell{l}{Deep-FSMN}} & Vanilla & 1/1 & 40.46 & 87.71 & 89.53 \\
& Distill & 1/1 & 40.46 & 90.02 & 90.95 \\
& {$\ell2$-norm} & {1/1} & {40.46} & {93.23} & {93.76} \\ 
& {FID} & 1/1 & 40.46 & 94.08 & 94.37 \\ 
& {FID + LPB} & {1/1} & {40.46} & {96.21} & {96.22} \\ 
\midrule
\multirow{12}{*}{\tabincell{l}{BiFSMNv2\\$_{\text{TBA}[1, 0.5, 0.25]\times}$}} & \multirow{3}{*}{Vanilla} & \multirow{3}{*}{1/1} & 35.01 & {95.09} & {95.30} \\
 & & & 27.96 & 94.75 & 95.12 \\
 & & & 24.43 & 92.99 & 93.58 \\
\cdashline{2-6}[1pt/1pt]
 & \multirow{3}{*}{Distill} & \multirow{3}{*}{1/1} & 35.01 & {95.48} & {95.57}  \\
 & & & 27.96 & 94.93 & 95.55 \\
 & & & 24.43 & 93.74 & 93.75 \\
\cdashline{2-6}[1pt/1pt]
 & \multirow{3}{*}{{$\ell2$-norm}} & {\multirow{3}{*}{1/1}} & {35.01} & {93.62} & {92.93}  \\
 & & & {27.96} & {56.80} & {60.89} \\
 & & & {24.43} & {56.80} & {60.89} \\
\cdashline{2-6}[1pt/1pt]
 & {\multirow{3}{*}{FID}} & \multirow{3}{*}{1/1} & 35.01 & 96.34 & 96.22  \\
 & & & 27.96 & 96.08 & 94.08 \\
 & & & \textbf{24.43} & 95.44 & 94.71 \\
\cdashline{2-6}[1pt/1pt]
 & {\multirow{3}{*}{FID + LPB (ours)}} & {\multirow{3}{*}{1/1}} & {35.01} & \textbf{{96.42}} & \textbf{{96.45}}  \\
 & & & {27.96} & {96.23} & {95.87} \\
 & & & {\textbf{24.43}} & {94.65} & {94.25} \\
\bottomrule
\end{tabular}}
\label{tab:ablation}
\end{table}

\begin{figure}[t]
\centering
\includegraphics[width=7cm]{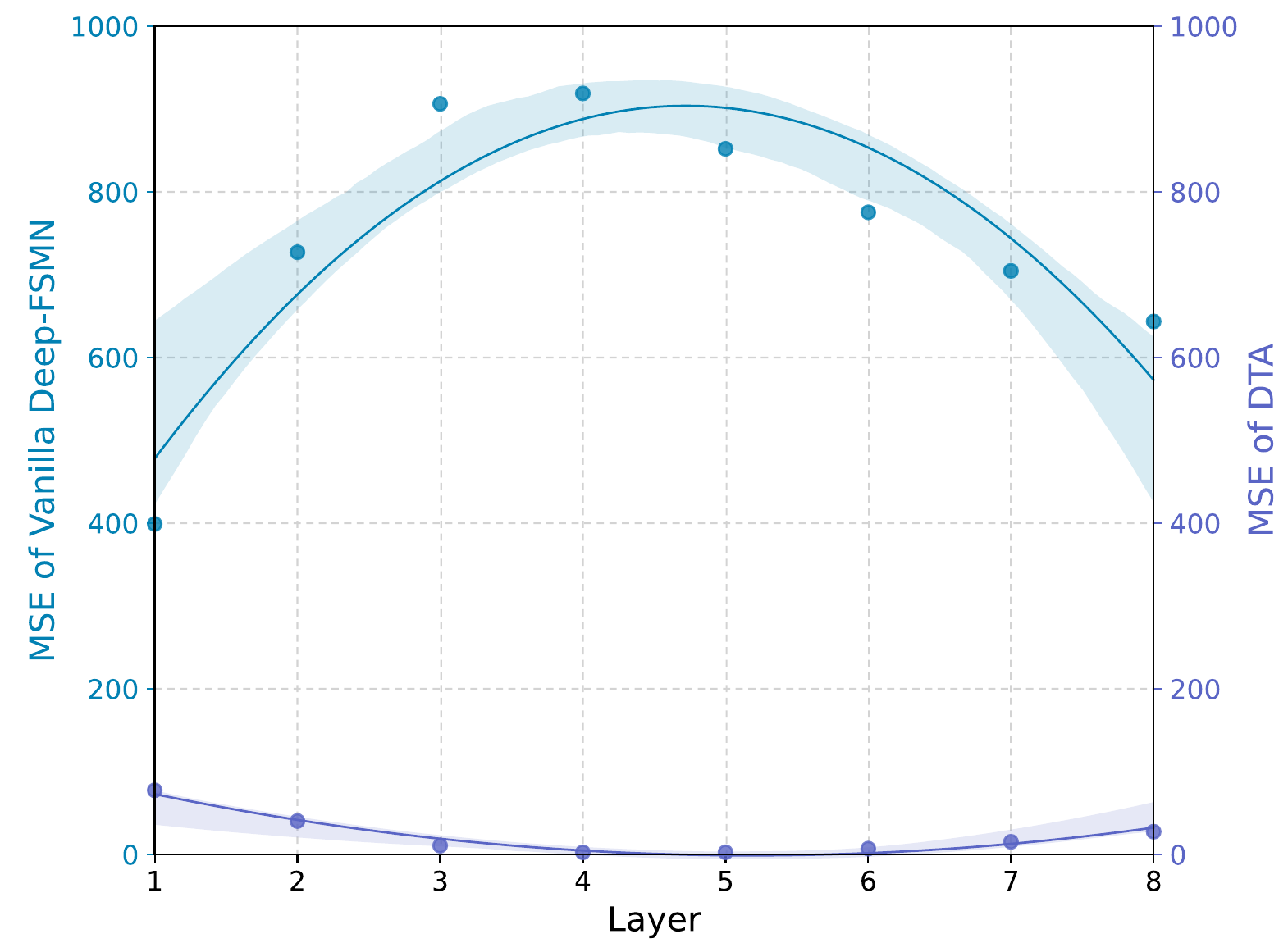}
\caption{{The comparison of activation quantization error measured by MSE between vanilla and DTA units on 8-layer Deep-FSMN architecture.}}
\label{fig:MSE-err}
\end{figure}

\begin{figure}[t]
\centering
\includegraphics[width=\linewidth]{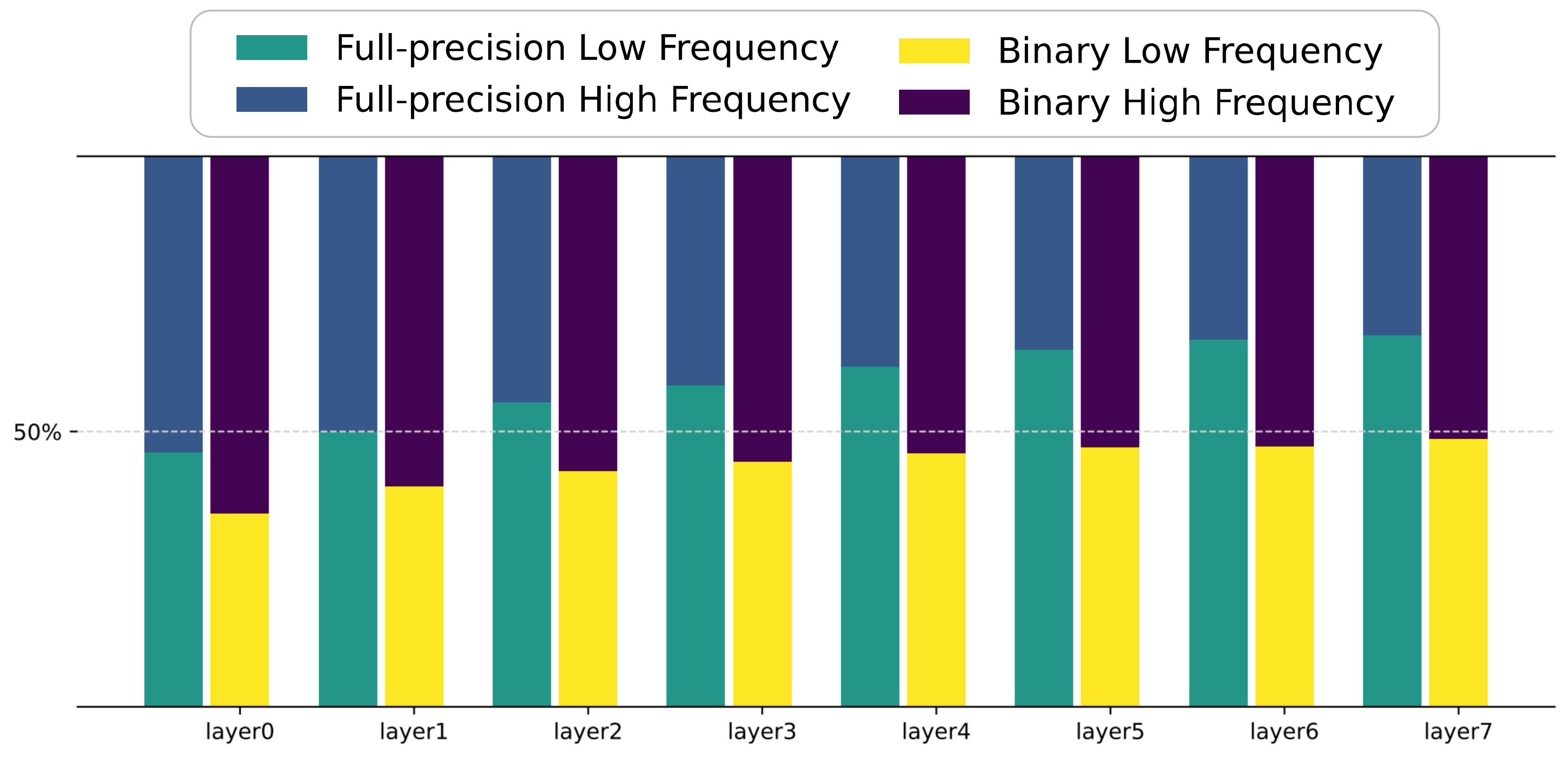}
\caption{{The percentage of the relative wavelet entropy of low/high frequency in each layer of full-precision and binarized models.}}
\label{fig:wavelet-bar}
\end{figure}

\begin{figure}[t]
\centering
\includegraphics[width=8cm]{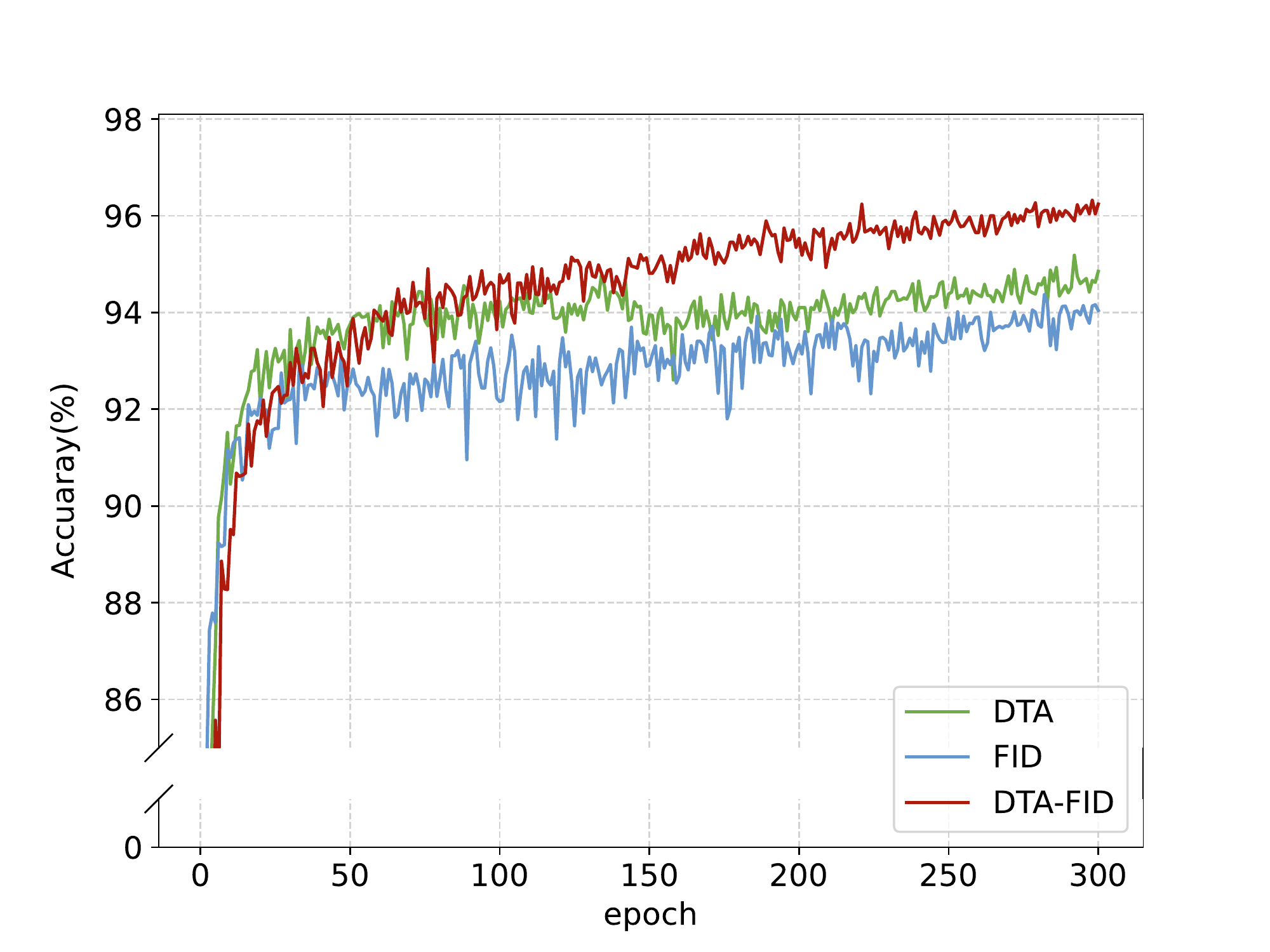}
\caption{The training curves for Deep-FSMN in FID, BiFSMNv2 in vanilla and BiFSMNv2 in FID.}
\label{fig:acc-log}
\end{figure}

\subsection{Ablation Study}

We conduct ablation experiments on the Google Speech Commands datasets (V1-12 tasks) to investigate the impact of the components of the proposed BiFSMNv2, including Dual-scale Thinnable 1-bit-Architecture (DTA) and Frequency Independent Distillation (FID).

\subsubsection{Effect of DTA} 
We first verify the effect of DTA. Compared with the traditional architecture, DTA significantly reduces the activation binarization error through two-scale binarization, so the information in the features can be recovered. {In Fig.~\ref{fig:MSE-err}, we randomly sample the activations in each layer in one forward and calculate the quantization error by comparing the binarized activation with the original floating-point ones measured by mean square error (MSE).} It shows the error in the corresponding layer in Deep-FSMN and DTA architectures, and we can see that DTA has significantly reduced binarization error. Additionally, the training curves in Fig. ~\ref{fig:acc-log} show that, compared to directly training the binarized version of Deep-FSMN, jointly training multiple versions of the DTA architecture is beneficial for model optimization and fast convergence. 

\subsubsection{Effect of FID}
FID is proposed to alleviate the frequency mismatch problem between binarized and full-precision representations in distillation. {In Fig.~\ref{fig:wavelet-bar}, we show the ratio of high and low frequencies between the input features of the FID full-precision and corresponding binarized layers.} The phenomenon in the figure shows that frequency mismatch is prevalent rather than accidental between full-precision and binarized models, so we use FID when distilling to align the frequencies in learning. The training curve in Fig.~\ref{fig:acc-log} also shows that BiFSMNv2 with FID converges more stable.

\subsubsection{{Effect of LPB}}
{Our LPB aims to further push the limits of binary neural networks. The ablation results demonstrate the strong effect of our LPB. Even though the application of DTA and FID improves the effect to a considerable height, LPB still brings a 0.23\% performance increase for BiFSMNv2 on Speech Commands V2-12 without any additional computational burden.}

\subsubsection{Accuracy Ablation Results}
As shown in Table~\ref{tab:ablation}, the vanilla binarized Deep-FSMN suffers a significant performance drop in both datasets. When solely utilizing DTA, the binarized computing units are strengthened, and the model is optimized by weighted losses in the backward propagation. Therefore, it shows a great deal of potential in not only the adaptive and lightweight computation at runtime but also the accuracy of the binarized network during inference. For example, using the vanilla and distillation optimization methods, our BiFSMNv2 with DTA surpasses Deep-FSMN by 7.38\% and 5.46\%, respectively.
And compared with the vanilla optimization, the results show that the application of the proposed FID further improves the distillation-based optimization by 4.06\% and 0.86\% compared to naive distillation on Deep-FSMN and DTA architecture, respectively.
{We also compare our FID technique with the distillation using a simple $\ell$2-norm. The experimental results show that it is not trivial to distill the binarized KWS networks directly. On both the Deep-FSMN and BiFSMNv2 architectures, FID has steadily improved the performance by about 3\% compared to the $\ell$2-norm distillation.}
It demonstrates that aligned frequency information makes the optimization of representations easier in binarization-aware distillation, proving that the distillation strategy is essential for aligning and transferring information. 
{The application of the proposed LPB pushes the BiFSMNv2 to a higher accuracy, and jointly using DTA, FID, and LBP further narrows the accuracy gap between the binarized model and the full-precision counterpart. The performance of 1-bit BiFSMNv2 on the Speech Commands V1-12 task achieves 96.42\%, just 1.51\% drop compared with the 32-bit counterpart.}

\begin{table}[!ht]
\centering
\caption{{Ablation for FLOPs of different parts between full-precision and binarized 8-layer Deep-FSMN architectures.}}
\setlength{\tabcolsep}{6.5mm}
{\begin{tabular}{llrr}
\toprule
{\textbf{Part}} & ~ & {\textbf{32-bit}} & {\textbf{1-bit}}  \\ \midrule
\multirow{2}{*}{{Head}} & {Layer1.1} & {14.08} & {14.08}  \\ 
~ & {Layer1.2} & {105.64} & {3.47}  \\ \midrule
{Neck} & {Layer2.1} & {33.55} & {1.18}  \\ \midrule
\multirow{8}{*}{{Backbone}} & {Block1} & {69.34} & {3.82}  \\ 
~ & {Block2} & {69.34} & {3.82}  \\ 
~ & {Block3} & {69.34} & {3.82}  \\ 
~ & {Block4} & {69.34} & {3.82}  \\ 
~ & {Block5} & {69.34} & {3.82}  \\ 
~ & {Block6} & {69.34} & {3.82}  \\ 
~ & {Block7} & {69.34} & {3.82}  \\ 
~ & {Block8} & {69.34} & {3.82}  \\ \midrule
{Classifier} & {Layer3.1} & {1.57} & {1.57} \\ \bottomrule
\end{tabular}}
\label{tab:flops}
\end{table}

\subsubsection{{Efficiency Ablation Results}}

{We present the ablation results for the FLOPs of the full-precision and binarized architecture to measure the acceleration effect of our proposed binarization. The results in Table~\ref{tab:flops} show that our binarization algorithm achieves a super-high speedup of 18.1$\times$ for a single block in backbone, while the FLOPs produced by the unbinarized layers are only a relatively small fraction of the full-precision counterpart. This guarantees the ultrahigh speedup of our BiFSMNv2.}

\subsection{Comparative Experiments}

{First, we compare our BiFSMNv2 with existing SOTA structure-independent binarization methods, including BNN~\cite{BNN}, DoReFa~\cite{dorefa}, XNOR~\cite{XNOR}, Bi-Real~\cite{BiReal}, IR-Net~\cite{qin2019forward}, RAD~\cite{Regularize-act-distribution}, ReCU~\cite{xu2021recu}, ReActNet~\cite{liu2020reactnet}, XNOR++~\cite{XNOR++}, LCR~\cite{shang2022lipschitz}, and FDA~\cite{xu2021learning}.} We evaluate these binarization methods on 8-block Deep-FSMN and BiFSMN with a similar size as the largest variant of 4-block BiFSMNv2. 

\begin{table}[!th]
\centering
\caption{{Comparison of SOTA binarization methods on Speech Commands V1 (SC-V1) and V2 (SC-V2) datasets with 12, 20, and 35 classifications (CLS 12, CLS 20, and CLS 35).}}
\setlength{\tabcolsep}{1.3mm}
{\begin{tabular}{llcrrrrr}
\toprule
{\bf Dataset} & {\bf Method}  & {\tabincell{c}{\textbf{\#Bits}\\$_\textrm{(W/A)}$}} & {\tabincell{r}{{\textbf{\#Para}}\\{$_\textrm{(M)}$}}}  & {\tabincell{r}{\textbf{FLOPs}\\$_\textrm{(M)}$}}  & {\tabincell{r}{{\textbf{Time}}\\{$_\textrm{(ms)}$}}}  & {\tabincell{r}{\textbf{SC-V1}\\$_{(\%)}$}} & {\tabincell{r}{\textbf{SC-V2}\\$_{(\%)}$}}\\
\midrule
\multirow{17}{*}{\tabincell{l}{CLS 12}} & Full Prec. & 32/32 &  {0.60} & 710.15 & {377} & 97.93 & 98.05\\
\cmidrule{2-8}
& DoReFa  & 1/1 & {0.60} & 40.46 & {81.9} & 66.42 & 66.59\\
& RAD     & 1/1 & {0.60} & 35.87 & {80.7} & 71.51 & 69.80\\
& BNN     & 1/1 & {0.60} & 35.87 & {80.7} & 68.84 & 70.87\\
& XNOR++  & 1/1 & {0.60} & 45.04 & {--} & 91.71 & 82.24\\
& IR-Net  & 1/1 & {0.60} & 40.46 & {81.9} & 86.81 & 85.10\\
& XNOR    & 1/1 & {0.60} & 45.04 & {--} & 82.74 & 87.34\\
& FDA     & 1/1 & {0.60} & 40.46 & {81.9} & 88.71 & 87.79\\
& Bi-Real & 1/1 & {0.60} & 40.46 & {81.9} & 85.87 & 87.93\\
& {LCR} & {1/1} & {0.60} & {40.46} & {81.9} & {93.14} & {93.51} \\
& ReCU    & 1/1 & {0.60} & 40.46 & {81.9} & 93.74 & 92.83\\
& ReActNet   & 1/1 & {0.60} & 45.96 & {81.9} & 93.96 & 94.02\\
\cdashline{2-8}[1pt/1pt]
& \multirow{3}{*}{\tabincell{l}{BiFSMN\\$_{[1, 0.5, 0.25]\times}$}}    & \multirow{3}{*}{1/1} & {\multirow{3}{*}{0.61}} & 40.46 & {81.9} & {95.03} & {94.86} \\
& & &  & 29.90 & {46.5} & 94.87 & 94.73\\
& & &  & {24.62} & {28.8} & 94.48 & 94.63\\
\cdashline{2-8}[1pt/1pt]
& \multirow{3}{*}{\tabincell{l}{BiFSMNv2\\$_{[1, 0.5, 0.25]\times}$}} & \multirow{3}{*}{1/1} & {\multirow{3}{*}{0.30}} & 35.01 & {58.4} & {\textbf{96.42}} & {\textbf{96.45}}\\
& & & & 27.96 & {34.7} & {96.23} & {95.87}\\
& & & & \textbf{24.43} & {22.9} & {94.65} & {94.25}\\
\midrule
\multirow{17}{*}{\tabincell{l}{CLS 20}} & Full Prec.    & 32/32 & {0.61} & 711.20 & {377} & 96.57 & 97.00\\
\cmidrule{2-8}
& IR-Net  & 1/1 & {0.61} & 41.50 & {81.9} & 83.78 & 83.32\\
& Bi-Real & 1/1 & {0.61} & 41.50 & {81.9} & 80.84 & 84.39\\
& XNOR    & 1/1 & {0.61} & 45.04 & {--} & 80.69 & 85.05\\
& FDA     & 1/1 & {0.61} & 41.50 & {81.9} & 83.10 & 87.80\\
& BNN     & 1/1 & {0.61} & 36.92 & {80.7} & 87.19 & 88.28\\
& ReCU    & 1/1 & {0.61} & 41.50 & {81.9} & 91.62 & 89.79\\
& XNOR++  & 1/1 & {0.61} & 46.09 & {--} & 93.20 & 89.74\\
& {LCR} & 1/1 & {0.61} & {41.50} & {81.9} & {90.21} & {92.26} \\
& DoReFa  & 1/1 & {0.61} & 41.50 & {81.9} & 90.02 & 91.27\\
& RAD     & 1/1 & {0.61} & 36.92 & {80.7} & 89.43 & 92.79\\
& ReActNet   & 1/1 & {0.61} & 47.01 & {81.9} & 90.35 & 92.18\\
\cdashline{2-8}[1pt/1pt]
& \multirow{3}{*}{\tabincell{l}{BiFSMN\\$_{[1, 0.5, 0.25]\times}$}}    & \multirow{3}{*}{1/1} & \multirow{3}{*}{{0.61}} & 41.50 & {81.9} & {92.88} & {92.98} \\
& & &  & 30.95 & {46.5} & 92.67 & 92.81\\
& & &  & {25.67} & {28.8} & 92.65 & 92.72\\
\cdashline{2-8}[1pt/1pt]
& \multirow{3}{*}{\tabincell{l}{BiFSMNv2\\$_{[1, 0.5, 0.25]\times}$}} & \multirow{3}{*}{1/1} & \multirow{3}{*}{{0.31}} & 36.06 & {58.4} & {\textbf{94.22}} & {\textbf{94.73}}\\
& & &  & 29.01 & {34.7} & {94.24} & {93.89}\\
& & &  & \textbf{25.48} & {22.9} & {92.48} & {92.58}\\
\midrule
\multirow{17}{*}{\tabincell{l}{CLS 35}} & Full Prec.    & 32/32 &  & 713.16 & {377} & 96.63 & 95.96 \\
\cmidrule{2-8}
& IR-Net  & 1/1 & {0.60} & 43.47 & {81.9} & 74.09 & 74.93\\
& FDA     & 1/1 & {0.60} & 43.47 & {81.9} & 85.73 & 79.61\\
& Bi-Real & 1/1 & {0.60} & 43.47 & {81.9} & 80.86 & 81.86\\
& XNOR    & 1/1 & {0.60} & 48.06 & {--} & 81.25 & 84.05\\
& XNOR++  & 1/1 & {0.60} & 48.06 & {--} & 85.03 & 90.23 \\
& BNN     & 1/1 & {0.60} & 38.88 & {80.7} & 85.93 & 84.13\\
& ReCU    & 1/1 & {0.60} & 43.47 & {81.9} & 89.45 & 84.99\\
& DoReFa  & 1/1 & {0.60} & 43.47 & {81.9} & 89.35 & 87.15\\
& ReActNet   & 1/1 & {0.60} & 48.98 & {81.9} & 90.08 & 88.96\\
& {LCR} & 1/1 & {0.60} & {43.47} & {81.9} & {90.17} & {88.55} \\
& RAD     & 1/1 & {0.60} & 38.88 & {80.7} & 91.32 & 89.94\\
\cdashline{2-8}[1pt/1pt]
& \multirow{3}{*}{\tabincell{l}{BiFSMN\\$_{[1, 0.5, 0.25]\times}$}} & \multirow{3}{*}{1/1} & \multirow{3}{*}{{0.63}} & 43.47 & {81.9} & {92.10} & {90.67}\\
& & &  & 32.91 & {46.5} & 91.93 & 90.54\\
& & &  & {27.63} & {28.8} & 91.85 & 90.42\\
\cdashline{2-8}[1pt/1pt]
& \multirow{3}{*}{\tabincell{l}{BiFSMNv2\\$_{[1, 0.5, 0.25]\times}$}} & \multirow{3}{*}{1/1} & \multirow{3}{*}{{0.32}} & 38.02 & {58.4} & {\textbf{94.57}} & {\textbf{93.05}}\\
& & &  & 30.97 & {34.7} & {94.47} & {92.04}\\
& & &  & \textbf{27.45} & {22.9} & {88.71} & {90.04}\\
\bottomrule
\end{tabular}}
\label{tab:methods}
\end{table}

\begin{table}[!th]
\centering
\caption{Comparison of various binarized architectures for KWS on Speech Commands V1-12 task.}
\resizebox{\linewidth}{!}
{\begin{tabular}{llcrrr}
\toprule
{\bf Arch.} & {\bf Quant}  & {\tabincell{c}{\textbf{\#Bits}$_\textrm{(W/A)}$}} & {\tabincell{r}{\textbf{\#Para}$_\textrm{(M)}$}}  & {\tabincell{r}{\textbf{FLOPs}$_\textrm{(M)}$}}  &{\bf \tabincell{r}{Acc.$_{(\%)}$}} \\
\midrule
\multirow{5}{*}{\tabincell{l}{VGG19bn}} & Full Prec.    & 32/32 & 38.97 & 53348.40 & 97.74\\
\cmidrule{2-6}
& IR-Net & 1/1 & 38.97 & 1030.18 & 63.86\\
& XNOR & 1/1 & 38.97 & 1061.63 & 66.10\\
& ReCU & 1/1 & 38.97 & 1030.18 & 56.45\\
\midrule
\multirow{5}{*}{\tabincell{l}{{BC-ResNet-4}}} & {Full Prec.} & {32/32} & {0.18} & {135.50} & {94.67}\\
\cmidrule{2-6}
& {XNOR} & {1/1} & {0.18} & {89.19} & {65.82}\\
& {IR-Net} & {1/1} & {0.18} & {78.35} & {56.10}\\
& {ReCU} & {1/1} & {0.18} & {78.35} & {66.19}\\
\midrule
\multirow{5}{*}{\tabincell{l}{{BC-ResNet-8}}} & Full Prec. & 32/32 & 0.35 & 3749.71 & 97.84\\
\cmidrule{2-6}
& XNOR & 1/1 & 0.35 & 619.03 & 65.94\\
& IR-Net & 1/1 & 0.35 & 541.70 & 66.51\\
& ReCU & 1/1 & 0.35 & 541.70 & 56.45\\
\midrule
\multirow{5}{*}{\tabincell{l}{{MHAtt-RNN}}} & {Full Prec.} & {32/32} & {0.86} & {13169} & {94.02}\\
\cmidrule{2-6}
& {XNOR} & {1/1} & {0.86} & {13124} & {56.45}\\
& {IR-Net} & {1/1} & {0.86} & {13123} & {65.53}\\
& {ReCU} & {1/1} & {0.86} & {13123} & {65.53}\\
\midrule
\multirow{5}{*}{\tabincell{l}{FSMN}} & Full Prec. & 32/32 & 0.45 & 1625.29 & 97.52\\
\cmidrule{2-6}
& XNOR & 1/1 & 0.45 & 80.61 & 55.89\\
& IR-Net & 1/1 & 0.45 & 64.36 & 88.35\\
& ReCU & 1/1 & 0.45 & 64.36 & 87.24\\
\midrule
\multirow{8}{*}{\tabincell{l}{Deep-FSMN}} & Full Prec. & 32/32 & 0.60 & 710.15 & 97.93\\
\cmidrule{2-6}
& XNOR & 1/1 & 0.60 & 45.04 & 82.74\\
& IR-Net & 1/1 & 0.60 & 40.46 & 86.81 \\
& ReCU & 1/1 & 0.60 & 40.46 & 93.74\\
& ReActNet & 1/1 & 0.60 & 45.96 & 93.96 \\
& HED & 1/1 & 0.60 & 40.46 & 93.47 \\
& \tabincell{l}{FID} & 1/1 & 0.60 & 40.46 & {94.08} \\
\midrule
\multirow{16}{*}{\tabincell{l}{BiFSMN\\$_{\text{TBA} [1, 0.5, 0.25]\times}$}}  & Full Prec. & 32/32 & 0.60 & 710.15 & 97.93 \\
\cmidrule{2-6}
& \multirow{3}{*}{\tabincell{l}{IR-Net}} & \multirow{3}{*}{1/1} & \multirow{3}{*}{0.61} & 40.46 & 88.37 \\
& & & & 29.90 & 87.11 \\
& & & & 24.62 & 86.08 \\
\cdashline{2-6}[1pt/1pt]
& \multirow{3}{*}{\tabincell{l}{ReCU}} & \multirow{3}{*}{1/1} & \multirow{3}{*}{0.61} & 40.46 & 92.45 \\
& & & & 29.90 & 92.16 \\
& & & & 24.62 & 91.85 \\
\cdashline{2-6}[1pt/1pt]
& \multirow{3}{*}{\tabincell{l}{ReActNet}} & \multirow{3}{*}{1/1} & \multirow{3}{*}{0.61} & 45.96 & 93.94 \\
& & & & 33.04 & 93.78 \\
& & & & 26.58 & 93.70 \\
\cdashline{2-6}[1pt/1pt]
 & \multirow{3}{*}{HED} & \multirow{3}{*}{1/1} & \multirow{3}{*}{0.61} & 40.46 & {95.03}\\
 & & & & 29.90 & 94.87\\
 & & & & {24.62} & 94.48\\
\cdashline{2-6}[1pt/1pt]
 & \multirow{3}{*}{FID} & \multirow{3}{*}{1/1} & \multirow{3}{*}{0.61} & 40.46 & {95.42}\\
 & & & & 29.90 & 95.02\\
 & & & & {24.62} & 94.56\\
\cdashline{1-6}[1pt/1pt]
\multirow{3}{*}{\tabincell{l}{BiFSMNv2\\$_{\text{TBA} [1, 0.5, 0.25]\times}$}} & {\multirow{3}{*}{FID + LPB}} & \multirow{3}{*}{1/1} & \multirow{3}{*}{\textbf{0.30}} & {{35.01}} & {\textbf{\textbf{96.42}}}\\
& & & & {27.96} & {{96.23}}\\
& & & & {\textbf{24.43}} & {{94.65}}\\
\midrule
\multirow{16}{*}{\tabincell{l}{BiFSMN$_\textrm{S}$\\$_{\text{TBA} [1, 0.5, 0.25]\times}$}}  & Full Prec. & 32/32 & 0.05 & 91.62 & 97.51 \\
\cmidrule{2-6}
& \multirow{3}{*}{\tabincell{l}{IR-Net}} & \multirow{3}{*}{1/1} & \multirow{3}{*}{0.05} & 11.94 & 72.21 \\ 
& & & & 10.08 & 71.70 \\ 
& & & & 9.16 & 71.66 \\ 
\cdashline{2-6}[1pt/1pt]
& \multirow{3}{*}{\tabincell{l}{ReCU}} & \multirow{3}{*}{1/1} & \multirow{3}{*}{0.61} & 11.94 & 89.54 \\
& & & & 10.08 & 89.20 \\
& & & & 9.16 & 88.33 \\
\cdashline{2-6}[1pt/1pt]
& \multirow{3}{*}{\tabincell{l}{ReActNet}} & \multirow{3}{*}{1/1} & \multirow{3}{*}{0.61} & 13.90 & 88.88 \\
& & & & 11.46 & 88.29 \\
& & & & 10.24 & 87.96 \\
\cdashline{2-6}[1pt/1pt]
& \multirow{3}{*}{HED} & \multirow{3}{*}{1/1} & \multirow{3}{*}{0.05} & 11.94 & {90.65}\\
 & & & & 10.08 & 90.33\\
 & & & & {9.16} & 90.31\\
\cdashline{2-6}[1pt/1pt]
 & \multirow{3}{*}{FID} & \multirow{3}{*}{1/1} & 
\multirow{3}{*}{0.05} & 11.94 & {90.71} \\
 & & & & 10.08 & 90.45 \\
 & & & & {9.16} & 90.33 \\
\cdashline{1-6}[1pt/1pt]
\multirow{3}{*}{\tabincell{l}{BiFSMNv2$_\textrm{S}$\\$_{\text{TBA} [1, 0.5, 0.25]\times}$}} & {\multirow{3}{*}{FID + LPB}} & \multirow{3}{*}{1/1} & \multirow{3}{*}{\textbf{0.03}} & {10.18} & {\textbf{92.02}}\\
& & & & {8.77} & {90.96}\\
& & & & {\textbf{8.06}} & {89.49}\\
\bottomrule
\end{tabular}}
\label{tab:architectures}
\end{table}

The results in Table~\ref{tab:methods} show that our 1-bit BiFSMNv2 completely outperforms existing SOTA binarization methods by a wide margin.
{On Speech Commands V1-12 tasks, BiFSMNv2 surpasses the classic XNOR and IR-Net 13.68\% and 9.61\%, respectively, and even outperforms the recent FDA and ReCU methods by 9.61\% and 2.68\%.
The advantages of BiFSMNv2 are also presented on V1-20 and V1-35 tasks and bring 3.87\% and 3.40\% accuracy improvement compared with the SOTA binarization methods, respectively, though these tasks are considered to be more challenging.}
Compared to the first version of BiFSMN, BiFSMNv2 wins by a great margin, including in the comparison of the accuracy of all thinner variants.
A similar phenomenon is on the Speech Commands V2 datasets. BiFSMNv2 surpasses SOTA binarization methods and BiFSMN on V2-12, V2-20, and V2-35 tasks.
{It is noteworthy that BiFSMNv2 even enjoys competitive accuracy to full-precision counterparts within a 1.60\% accuracy drop on both datasets. \textit{e.g.}, it only drops 1.51\% accuracy in Speech Command V1-12 task. 
Furthermore, from the perspective of inference cost, the thinner version BiFSMNv2$_{0.5\times}$ with two blocks and BiFSMNv2$_{0.25\times}$ with one block achieves an amazing 25.40$\times$ and 29.07$\times$ FLOPs saving without sacrificing accuracy (only 0.19\% and 1.77\% drop) on Speech Commands V1-12 task.}

{To further validate the advantage of our BiFSMNv2 from the neural architecture perspective, we also compare it with various networks widely used in KWS, including FSMN~\cite{zhang2015feedforward}, VGG19bn~\cite{VeryDeepConvolutional} (selected from the open-source project~\cite{tugstugi2023} for Speech Commands KWS tasks), BC-ResNet-4/BC-ResNet-8~\cite{kim2021broadcasted}, MHAtt-RNN~\cite{rybakov2020streaming}, and BiFSMN (the conference version)~\cite{qin2022bifsmn}.}
We binarized these architectures with four representative binarization methods, XNOR, IR-Net, ReActNet, and ReCU. In Table~\ref{tab:architectures}, our FID can generally be applied in Deep-FSMN-based architectures, including Deep-FSMN, BiFSMN, and BiFSMNv2and improves their performance. 
{The results show that our BiFSMNv2 is with a significant advantage compared to existing binarized architectures. Except for the Deep-FSMN with a similar number of parameters, BiFSMNv2 also far surpasses the binarized VGG19bn by 30.32\% on the V1-12 task, which has several thousand times of parameters.
BiFSMNv2 has also achieved significant advantages compared to the binarized versions of the architectures specially designed for KWS. Specifically, compared to binarized BC-ResNet, BiFSMNv2 can achieve up to 40.32\% advantage, and it can even exceed at least 29.91\% compared to the binarized BC-ResNet-8 with more parameters and higher FLOPs. The binarization of MHAtt-RNN also causes a severe accuracy drop, which shows that it is difficult and risky to binarize the existing KWS architecture directly, and shows the significant advantages of our BiFSMNv2 compared to the existing architecture of binarization.}
We further slim the model width and provide an extremely tiny BiFSMNv2$_\textrm{S}$ (kernel size is 3$\times$3, and the number of channels is 12 in the front-end, 32 backbone memory layers and 56 in hidden layers, respectively) with only 0.03M parameters and 8.06$\sim$10.18M FLOPs. The results show that our BiFSMNv2$_\textrm{S}$ still surpasses the BiFSMN$_\textrm{S}$ though the latter has heavier storage and computation, demonstrating that our techniques also work well on tiny binarized networks.

{In addition, we compare the parameter amount and time delay of different binarization methods under the same deployment environment in Table~\ref{tab:methods}. We use daBNN~\cite{zhang2019dabnn}, one of the most widely used binarization open-source deployment frameworks, to deploy different architectures on real edge hardware and evaluate the inference speed (``-" indicates the binarization method cannot be supported). The results show that our BiFSMNv2 outperforms other methods by a large margin both in parameter amount and time delay. Because the structure can be thinned and has fewer FLOPs, BiFSMNv2 has the least delay, and his 0.25$\times$ variant can even have an inference delay of only 22.9ms, which is 57.8ms less than BNN, and less than the 0.25 variant structure of BiFSMN 5.9ms. And due to fewer blocks, the parameter number of BiFSMNv2 is less to 0.32M, only 1/2 of other methods.}

\begin{figure*}[t]
\centering
\includegraphics[width=\linewidth]{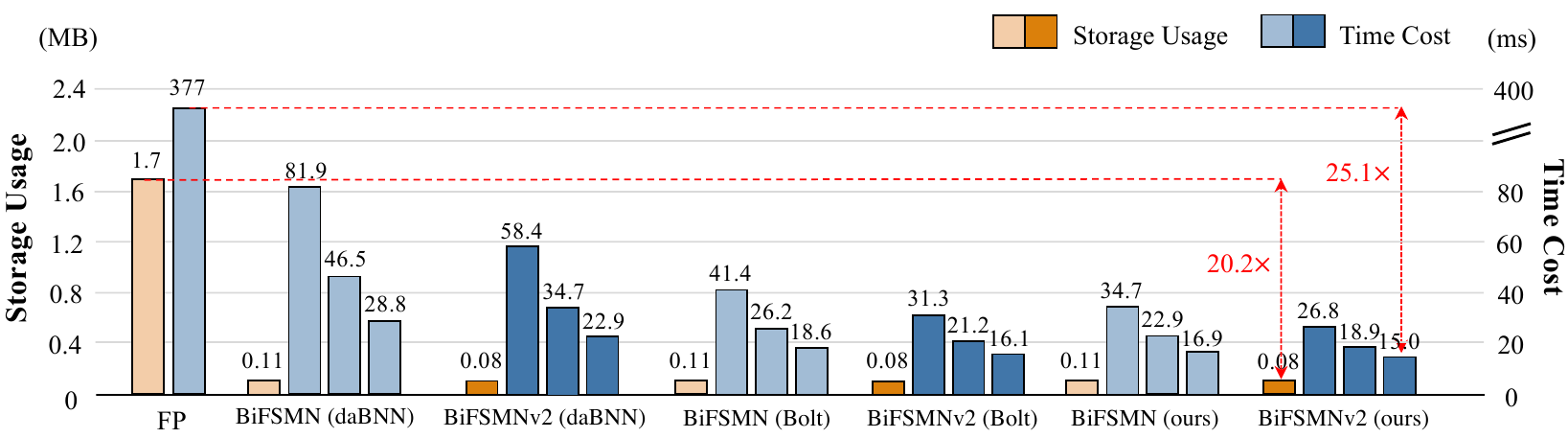}
\caption{Evaluation of inference performance on real-world ARMv8 devices.}
\label{fig:arm_performance}
\end{figure*}

\subsection{Deployment Efficiency}

As mentioned, low memory footprint and fast real-time response are highly expected in KWS while running on real-world edge devices. 

According to Fig.~\ref{fig:arm_performance}, due to the proposed optimized 1-bit Fast Bitwise Computation Kernel, our BiFSMNv2 delivers 20.2$\times$ storage saving compared to the full-precision counterparts.
Furthermore, benefiting from the thinnable architecture, BiFSMN can adaptively balance accuracy and efficiency at runtime according to the resources on the device and switches to BiFSMNv2$_{0.5\times}$ (2 blocks) or BiFSMNv2$_{0.25\times}$ (1 blocks) for further 19.9$\times$ and 25.1$\times$ speedups, respectively. It is much faster than the existing open-source, high-performance binarization frameworks (daBNN and Bolt) and shows that our BiFSMNv2 can satisfy different resource constraints.

\section{Conclusion}
We propose BiFSMNv2, an accurate and highly efficient binary neural network for KWS. We first construct a DTA to recover the representation capability of the binarized computation units by dual-scale activation binarization and liberate the speedup potential from an overall architecture perspective. We also construct an FID scheme to distill the high and low-frequency components independently to mitigate the information mismatch between full-precision and binarized representations. 
BiFSMNv2 outperforms existing binarization methods by convincing margins and is comparable to its full-precision counterparts. Furthermore, our ARMv8 real-world device with BiFSMNv2 implementation achieves an impressive 25.1$\times$ speedup and 20.2$\times$ storage-saving with the help of the proposed FBCK.
Our work demonstrates the great potential of binarization for KWS on resource-limited hardware and first pushes it to real-network performance. We hope our work will inspire future research on lightweight KWS.

\section{Acknowledgement}

This work was supported by The National Key Research and Development Plan of China (2020AAA0103503) and the Academic Excellence Foundation of BUAA for PhD Students.

\clearpage

{
\bibliographystyle{plain}
\bibliography{tnnls}
}

\end{document}